%% file: paper.tex
\documentclass[nohyperref]{article}

\usepackage[notheorems]{eda} 

\usepackage{cite}
\usepackage{amsmath,amssymb,amsfonts}
\usepackage{graphicx}
\usepackage{textcomp}
\usepackage{xcolor}
\usepackage{latexsym}
\usepackage{subcaption}
\usepackage{cancel} %

\usepackage{microtype}

\usepackage{hyperref}

\usepackage[accepted]{icml2022}

\usepackage{amsthm}

\usepackage[capitalize,noabbrev]{cleveref}

\theoremstyle{plain}

\theoremstyle{definition}

\theoremstyle{remark}

\usepackage[textsize=tiny]{todonotes}

\graphicspath{ {./figs/,./cached_figs/} }

\ifpdf
\usetikzlibrary{external}
\fi

\usetikzlibrary{patterns,positioning}

\usepackage{pifont}
\newcommand{\cmark}{\textcolor{applegreen}{\ding{51}}}
\newcommand{\xmark}{\textcolor{red}{\ding{55}}}

\include{defines}

\newcommand{\ourstyledtitle}{Label-Descriptive Patterns and Their Application to \\Characterizing Classification Errors}

\newcommand{\shorttitle}{Label-Descriptive Patterns  and Their Application to Characterizing Classification Errors}
\newcommand{\ourmethod}{\textsc{Premise}\xspace}
\newcommand{\oururl}{\url{https://github.com/uds-lsv/premise}}
\newcommand{\codeurl}{\oururl}

\icmltitlerunning{\shorttitle}

\begin{document}

\twocolumn[
\icmltitle{\ourstyledtitle}

\icmlsetsymbol{equal}{*}

\begin{icmlauthorlist}
\icmlauthor{Michael A. Hedderich}{equal,su}
\icmlauthor{Jonas Fischer}{equal,mpi}
\icmlauthor{Dietrich Klakow}{su}
\icmlauthor{Jilles Vreeken}{cis}
\end{icmlauthorlist}

\icmlaffiliation{su}{Saarland University, Saarland Informatics Campus, Saarbr\"ucken, Germany.}
\icmlaffiliation{mpi}{Max Planck Institute for Informatics, Saarbr\"ucken, Germany.}
\icmlaffiliation{cis}{CISPA  Helmholtz  Center  for  Information  Security, Saarbr\"ucken, Germany}

\icmlcorrespondingauthor{Michael A. Hedderich}{mhedderich@lsv.uni-saarland.de}
\icmlcorrespondingauthor{Jonas Fischer}{fischer@mpi-inf.mpg.de}

\vskip 0.3in
]

\printAffiliationsAndNotice{\icmlEqualContribution} %

\begin{abstract}
\input{abstract}

\end{abstract}

\input{intro.tex}

\input{related.tex}

\input{preliminaries.tex}

\input{theory.tex}

\input{approach.tex}

\input{experiments.tex}

\input{conclusion.tex}

\section*{Acknowledgment}
This work is partially funded by the Deutsche Forschungsgemeinschaft (DFG, German Research Foundation) – Project-ID 232722074 - SFB 1102.

\bibliography{bib/abbreviations,bib/bib-jilles,bib/bib-paper}
\bibliographystyle{icml2022}

\newpage
\appendix

\include{appendix}

\end{document}

%% file: defines.tex
\newcommand{\Normal}{\ensuremath{\mathcal{N}}}
\newcommand{\grab}{\textsc{Grab}}
\newcommand{\ripper}{\textsc{Ripper}\xspace}

\newcommand{\cortana}{\textsc{Cortana}\xspace}
\newcommand{\Data}{\ensuremath{D}}

\newcommand{\Transaction}{\ensuremath{t}}
\newcommand{\Transactions}{\ensuremath{T}}

\newcommand{\Item}{\ensuremath{I}}
\newcommand{\Items}{\ensuremath{\mathcal{\Item}}\xspace}
\newcommand{\Model}{\ensuremath{M}\xspace}
\newcommand{\Models}{\ensuremath{\mathcal{\Model}}\xspace}

\newcommand{\atoms}{\ensuremath{\mathit{it}}\xspace}

\newcommand{\X}{\ensuremath{X}}
\newcommand{\Pattern}{\ensuremath{P}\xspace}
\newcommand{\Patterns}{\ensuremath{\mathcal{P}}\xspace}
\newcommand{\lp}{\ensuremath{l_{+}}\xspace}
\newcommand{\lm}{\ensuremath{l_{-}}\xspace}
\newcommand{\Dll}{\ensuremath{D^l}\xspace}
\newcommand{\Dlp}{\ensuremath{D^{+}}\xspace}
\newcommand{\Dlm}{\ensuremath{D^{-}}\xspace}

\newcommand{\cl}{\ensuremath{\mathit{L}}\xspace}
\newcommand{\cld}{\ensuremath{\cl}}
\newcommand{\Clause}{\ensuremath{\mathit{cl}}}
\newcommand{\Embedding}{\ensuremath{\mathit{emb}}}
\DeclareRobustCommand{\stirling}{\genfrac\{\}{0pt}{}}
\newcommand{\Candidates}{\ensuremath{C}}
\newcommand{\Neighbor}{\ensuremath{\mathit{nb}}}

\DeclareRobustCommand{\symdiff}{\Delta}

\newcommand{\Projection}[2]{\ensuremath{\pi_{#1}\left(#2\right)}}
\newcommand{\Selection}[2]{\ensuremath{\sigma_{#1}\left(#2\right)}}

\newcommand{\Spumante}{\textsc{SPuManTe}\xspace}
\newcommand{\Grab}{\textsc{Grab}\xspace}
\newcommand{\Classy}{\textsc{Classy}\xspace}
\newcommand{\Anchors}{\textsc{Anchors}\xspace}
\newcommand{\Subgroup}{\textsc{Subgroup-Discovery}\xspace}
\newcommand{\Tree}{\textsc{Tree}\xspace}
\newcommand{\CSalt}{\textsc{C-SALT}\xspace}

\makeatletter
\newcommand\incircbin
{%
  \mathpalette\@incircbin
}
\newcommand\@incircbin[2]
{%
  \mathbin%
  {%
    \ooalign{\hidewidth$#1#2$\hidewidth\crcr$#1\bigcirc$}%
  }%
}
\DeclareRobustCommand{\oland}{\incircbin{\land}}
\DeclareRobustCommand{\oxor}{\incircbin{\times}}

\newcommand{\OLAND}[1]{$\oland$(\textit{#1})}
\newcommand{\OLOR}[1]{\normalfont$\oxor$(\textit{#1})}

\DeclareMathOperator*{\argmax}{argmax}

\makeatother

\pgfkeys{/pgfplots/tuftelike/.style={
  semithick,
  tick style={major tick length=4pt,semithick,black},
  separate axis lines,
  axis x line*=bottom,
  axis x line shift=10pt,
  xlabel shift=5pt,
  axis y line*=left,
  axis y line shift=10pt,
  ylabel shift=5pt}}

\def\cl{\mathit{L}}

\def\naturalsNoZero{\mathbb{N}}

\newcommand{\logbinom}[2]{\log \binom{#1}{#2}}

\newcommand{\LN}{\cl_\naturalsNoZero}
\newcommand{\LPC}{\cl_{pc}}
\pgfplotsset{compat=newest}

%% file: abstract.tex
State-of-the-art deep learning methods achieve human-like performance on many tasks, but make errors nevertheless. Characterizing these errors in easily interpretable terms gives insight into whether a classifier is prone to making systematic errors, but also gives a way to act and improve the classifier. We propose to discover those feature-value combinations (i.e., patterns) that strongly correlate with correct resp. erroneous predictions to obtain a global and interpretable description for arbitrary classifiers. We show this is an instance of the more general label description problem, which we formulate in terms of the Minimum Description Length principle. To discover a good pattern set, we develop the efficient \ourmethod algorithm. Through an extensive set of experiments we show it performs very well in practice on both synthetic and real-world data. Unlike existing solutions, it ably recovers ground truth patterns, even on highly imbalanced data over many features. Through two case studies on Visual Question Answering and Named Entity Recognition, we confirm that \ourmethod gives clear and actionable insight into the systematic errors made by modern NLP classifiers.

%% file: intro.tex
\section{Introduction}

State-of-the-art deep learning methods are known for their ability to achieve human-like performance on challenging tasks. As much as `to err is human,' these classifiers make errors too. Some of these errors are due to noise that is inherent to the process we want to model, and therewith relatively benign. Systematic errors, on the other hand, e.g. those due to bias or misspecification, are much more serious as these lead to models that are inherently unreliable. If we know under what conditions a model performs poorly, we can actively intervene, e.g., by augmenting the training data, and so improve overall reliability and performance. Before we can do so, we first need to know whether a classifier makes systematic errors, and if so, how to characterize them in easily understandable terms. 

Given a dataset with labels that specify which instances were classified correctly or incorrectly, we are interested in finding combinations of features that describe where the classifier's predictions are incorrect. For a Natural Language Processing (NLP) task, the input features are words. If, for example, the combination of words ``\textit{how, many}'' is primarily found in misclassified instances, this can indicate that our classifier struggles with the concept of counting. A toy example is visualized in Figure \ref{fig:toy_nlp_example}.

\begin{figure}
	\centering
	
	\begin{tabular}{l c}
		\toprule
		\multicolumn{2}{l}{Instances \hspace{2.8cm} Correct Prediction?} \\ \midrule
		\textbf{How many} ducks are in the picture? & \xmark \\
		What are the ducks eating? & \xmark \\
		\textbf{How many} roosters are in the puddle? & \xmark \\
		Do you see ducks in the puddle? & \cmark \\ 
		Are there many ducks playing? & \cmark \\ 
		\bottomrule
	\end{tabular}
	\caption{Toy example with input instances and the label specifying if the classifier predicted correctly. The pattern \OLAND{how, many} correlates with misclassification. The word \textit{ducks} is also a frequent pattern but independent of the label and therefore not of relevance.\label{fig:toy_nlp_example}} 
\end{figure}

Local explanation methods such as LIME \cite{local/ribeiro2016LIME} describe the decision boundary of each instance. In contrast, we are interested in an efficient way to obtain a \emph{global} and \emph{non-redundant} description of our classifier's issues on the given input data. To this end, we turn to pattern mining. Here, a combination of features is a pattern, and we look for the set of patterns that best characterizes on which instances the classifier tends to perform poorly. This can be phrased as the more general problem of label description. For data with binary features, we are interested in the associations between the feature data and the labels. We formulate this problem in terms of the Minimum Description Length (MDL) principle, which identifies the best set of patterns as the one that best compresses the data without loss. 

To capture phenomena of text input, e.g., synonyms, we consider a rich pattern language that allows us to express conjunctions, mutual exclusivity, and nested combinations thereof. As the search space is twice exponential and does not exhibit any easy-to-exploit structure, we propose the efficient and hyper-parameter-free \ourmethod algorithm to heuristically discover the \emph{premises} under which we see the given predictions.

We evaluate \ourmethod both on synthetic and real-world data. We show that, unlike the state of the art in data mining, \ourmethod is robust to noise, scales to large numbers of features, and deals well with class imbalance, as well as varying association strength of patterns to labels. Through two case studies, we show that \ourmethod discovers patterns that provide clear insight into the systematic errors of NLP classifiers. For Visual Question Answering (VQA), we elucidate the issues of two classifiers~\cite{vqa/Zhu2016Visual7W,vqa/Tan2019LXMERT}, including aspects like counting, spatial orientation, and higher reasoning. For a neural Named Entity Recognition (NER) model~\cite{ner/ma2016lstmcnncrf}, we show that \ourmethod discovers patterns that are both interpretable and that can be acted upon.

%% file: related.tex
\section{Related Work}
\label{sec:rel}

\subsection*{Label Description in Data Mining}

Describing labels in terms of features is obviously related to classification. Here, however, we are not so much interested in prediction, but rather description and therewith value interpretability of the results over accuracy. We share this notion with emerging pattern mining~\cite{dong:99:efficient} which aims to discover those conditions under which a target attribute has an exceptional distribution. The key difference is that we are not interested in discovering \emph{all} patterns that are associated, which would be overly redundant and hard to interpret as a whole, but rather want a small and non-redundant set of patterns capturing relevant associations. 

Subgroup discovery~\cite{wrobel1997subgroup,subgroup/garcia2018survey} returns the top-$k$ patterns that correlate most strongly. This keeps the result sets of manageable size but does not solve the problem of redundancy~\cite{leeuwen:12:dssd}. Statistical pattern mining aims to discover patterns that correlate \emph{significantly} to a class label \cite{webb:07:discovering, lopez:15:wyl, papaxanthos:16:significant, pellegrina:18:significant}. In practice, these methods discover many hundreds of thousands of `significant' patterns even for small data. For surveys we refer to~\cite{ novak:09:sdr,subgroup/atzmueller2015survey,subgroup/garcia2018survey}.

Rule mining aims to discover rules of the form $X\rightarrow Y$~\cite{agrawal:93:assoc, hamalainen:12:kingfisher}, lending themselves to describe labels, too. Like above, most existing methods evaluate patterns individually, thereby discovering millions of rules even if the data is pure noise.
\Grab~\cite{fischer:19:grab} instead mines small sets of rules that together summarize the data well, an approach that is strongly related to ours.
One drawback for our setting, however, is that a rule found by \Grab does not make a statement about the data where the rule does \textit{not} apply. Here, we are interested
in exactly this differential description: where does the consequent appear predominantly in one label and not the other.
\Classy~\cite{proenca2020classy} instead discovers rule lists that specifically characterize a given label.
We show that both these approaches do not scale well and are sensitive to label imbalance.

In pattern mining, the goal is instead to describe binary data in terms of relevant feature co-occurrences. Specific works based on boolean matrix factorization~\cite{miettinen:12:joint} or the minimum description length principle~\cite{budhathoki:15:diffnorm} addressed the problem of finding patterns that are common and distinct between multiple databases.
While this can be applied in our setting, where labels induce a partition of the data into multiple databases, these algorithms do not scale to the applications that we consider.
The boolean matrix factorization approach \CSalt~\cite{hess17CSALT} was specifically designed for data with given class labels. Our results show, however, that it has low recall, missing most interesting patterns.

\subsection*{Explainable ML and Misclassification}

Specifically to explain classifiers, several proposed approaches aim to capture dependencies of features or attributes that a classifier uses to make a prediction, e.g., in terms of patterns or rules~\cite{henelius2014peek, barakat2005eclectic}, by model distillation~\cite{frosst2017distilling, lakkaraju2017interpretable}, or to discover patterns of neurons within neural networks that drive a decision~\cite{fischer2021explainn}.
These, however, focus on the dependencies the classifier exploits for successful prediction as opposed to understanding where -- or why -- something goes wrong.
Here, \citet{wouter2014scape} use the \cortana tool \cite{Meeng2011Cortana} to explain where a classifier performs particularly poorly in terms of feature subspaces. SliceFinder~\cite{related/Chung2020Slicer} follows a similar idea. However, both models were only evaluated on data with less than 50 features. Our experiments show that these methods do not scale well to the feature spaces common in NLP data. 

Among the first methods of rule mining in NLP are template-based information extraction approaches~\cite{elaine:99:rapier, riloff:03:bootstrap}. These distantly related methods require pre-specified templates for which then rules that fulfill the templates can be extracted from given text. 
More recently, manual approaches based on challenging test sets~\cite{related/gardner2020contrast, related/Ribeiro2020Beyond} or testing a hypothetical cause for misclassification~\cite{related/Rondeau2018Systematic, related/Wu2019Errudite, related/Lee2019QADiver} have been suggested.
Such manual approaches, however, require existing knowledge about the difficulties of the models.
Local explanation methods such as LIME~\cite{local/ribeiro2016LIME} provide insights into what changes in the input influence a classifier's decision on a specific instance. \Anchors~\cite{local/ribeiro18anchor} obtains such explanations in an interpretable form similar to our patterns. As they need to explore the local decision boundary, they require, however, multiple classifier evaluations per instance. For a survey focused on local methods for explainable NLP, we refer to \citet{xai/Danilevsky20Survey}.

Here, we propose to mine sets of patterns that provide concise, interpretable, and global descriptions of the given label, which we formulate in MDL terms.
We further propose an efficient heuristic to discover such pattern sets in practice, which we test against state-of-the-art across all aforementioned fields on synthethic data with known ground truth, as well as real world case studies.
We show that \ourmethod is the only approach to be scalable and robust to noise and label imbalance while retrieving succinct pattern sets, all of which is crucial to tackle real world applications.

%% file: preliminaries.tex
\section{Preliminaries}

In this section, we introduce the notation we use throughout the paper and give a brief primer to MDL.  

\subsection{Notation}

We consider binary data, such as a sequence of input words of an NLP task where each word of the vocabulary is a binary feature (bag-of-words, word is present or not present). In data mining terms, each instance of our dataset is a transaction and each word present in the instance is an item of the transaction. For each instance, we also have a label that expresses whether the instance is misclassified by our classifier. Our whole dataset can then be described as binary transaction data \Data\ over a set of items \Items, where each transaction $t\in \Data$ is assigned a binary label $\ell(t)\in\{\lm,\lp\}$.
For ease of notation, we define the partition of the database according to this binary label $\Dlm = \{t\in \Data\mid \ell(t)=\lm\}$  and $\Dlp = \{t\in \Data\mid \ell(t)=\lp\}$.
In general, $\X \subseteq \Items$ denotes an itemset, the set of transactions that contain $X$ is  defined as $\Transactions_\X = \{ \Transaction \in \Data \mid \X \subseteq \Transaction \}$.
The projection of $\Data$ on an itemset $\X$ is $\Projection{\X}{\Data} = \{\Transaction \cap \X \mid \Transaction \in \Data \}$.

We are looking for human-interpretable associations of items that best explain a given database partition.
We describe these associations in terms of 'patterns', which we define by logical conditions over sets of items.
For a logical condition $c$, we define a selection operator as
$\Selection{c}{\Data} = \{\Transaction \in \Data \mid c(\Transaction) \equiv \top \}.$
 For an item $\Item \in \Items$, it holds that $\left[ c_\Item(\Transaction)  \equiv \top \leftrightarrow \Item \in \Transaction \right]$.
The $k$-ary AND operator $\oland(c_1,\ldots,c_k)$ describes patterns of co-occurrence and holds iff all its conditions hold.
Similarly, the $k$-ary XOR operator $\oxor$ describes patterns of mutual exclusivity and holds if exactly one of its condition holds.
We denote $\atoms(c)$ for the items in the condition and define the projection on a condition as $\Projection{c}{\Data} = \Projection{\atoms(c)}{\Data}$. Conditions can be nested; specifically we are interested in patterns of AND operator over XOR operations, i.e. $\oland(\oxor_{c_1,\ldots,c_k},\ldots,\oxor_{c'_1,\ldots,c'_k})(t)$. An XOR operation is called clause, $\gamma(c)$ lists all clauses in conjunctive condition $c$. To simplify notation, we drop $\Transaction$ when clear from context, write $\Item$ for conditions on a single item $c(\Item)$, and use condition and pattern  interchangeably.

\subsection{Minimum Description Length}
The Minimum Description Length (MDL) principle \cite{rissanen:78:mdl} is a practical approximation of Kolmogorov complexity \cite{vitanyi:93:book} that is both statistically well-founded and computable. It identifies the best model $\Model^*$ for data \Data\ out of a class of models \Models as the one that obtains the maximal lossless compression. For refined, or one-part, MDL, the length of the encoding in bits is obtained using the entire model class $\cl(\Data\mid\Models)$. While this variant of MDL provides strong optimiality guarantess \cite{grunwald:07:book}, it is only attainable for certain model classes. In practice, crude two-part MDL is often used, which computes the length of the model encoding $\cl(\Model)$ and the length of the description of the data given the model $\cl(\Data\mid\Model)$ separately. The total length of the encoding is then given as $\cl(\Model) + \cl(\Data\mid\Model)$. We use one-part MDL where possible and two-part MDL otherwise. Here, we are only interested in the codelengths and not the actual codes. Codelengths are measured in bits, hence all log operations are base 2 and we define $0 \log 0 = 0$.

%% file: theory.tex
\section{Theory}
\label{sec:theory}

To discover those patterns best describing the given labels, we first provide an intuition behind the concepts of the Minimum Description Length (MDL) principle in terms of our problem, and then formally introduce  the class of models $\mathcal{M}$ and corresponding codelength functions.

\subsection{The Problem, informally}

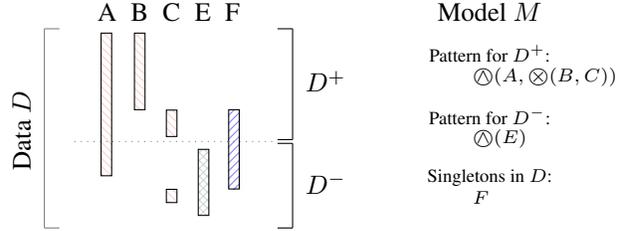
\begin{figure}[t]
\tikzsetnextfilename{toy-ex-premise}
\centering
 \begin{tikzpicture}[level/.style={level distance=10mm}]
 \tikzstyle{level 1}=[sibling distance=9mm]
 \tikzstyle{level 2}=[sibling distance=5mm]

  \node [anchor=south] (data) at (-1.6,-1.55) {\rotatebox{90}{Data $\Data$}};

 \draw[color=black!50] (-1.3,0.45) -- (-1.1,0.45);
 \draw[color=black!50] (-1.3,-2.2) -- (-1.3,0.45);
 \draw[color=black!50] (-1.3,-2.2) -- (-1.1,-2.2);

 \draw [dotted, thin, color=black!50] (-0.9,-1.05) -- (1.8,-1.05);

 \draw[color=black] (2,0.45) -- (1.8,0.45);
 \draw[color=black] (2,0.45) -- (2,-1.025);
 \draw[color=black] (2,-1.025) -- (1.8,-1.025);
 \node [anchor=west] (data) at (2.05,-0.25) {$\Dlp$};
 
 \draw[color=black] (2,-1.075) -- (1.8,-1.075);
 \draw[color=black] (2,-1.075) -- (2,-2.2);
 \draw[color=black] (2,-2.2) -- (1.8,-2.2);
 \node [anchor=west] (data) at (2.05,-1.6) {$\Dlm$};

 \node [anchor=north] (l1) at (-0.9,0.8) {};
 \node [anchor=east, right=5pt of l1, inner sep = 0pt, minimum width = 0pt] (l1A) {A}; 
 \node [anchor=west, right=5pt of l1A, inner sep = 0pt, minimum width = 0pt] (l1B) {B}; 
 \node [anchor=west, right=5pt of l1B, inner sep = 0pt, minimum width = 0pt] (l1C) {C};
 \node [anchor=west, right=5pt of l1C, inner sep = 0pt, minimum width = 0pt] (l1D) {E}; %
 \node [anchor=west, right=5pt of l1D, inner sep = 0pt, minimum width = 0pt] (l1E) {F};

 \node [anchor = west, below right=1pt and -2pt of l1A.south west] (l1Atl) {};
 \node [anchor = west, below right=55pt and 2pt of l1A.south west] (l1Abr) {};
 \draw [pattern=north west lines, pattern color=red!20] (l1Atl) rectangle (l1Abr);
 
 \node [anchor = west, below right=1pt and -2pt of l1B.south west] (l1Btl) {};
 \node [anchor = west, below right=30pt and 2pt of l1B.south west] (l1Bbr) {};
 \draw [pattern=north west lines, pattern color=red!20] (l1Btl) rectangle (l1Bbr);

 \node [anchor = west, below right=30pt and -2pt of l1C.south west] (l1Ctl) {};
 \node [anchor = west, below right=40pt and 2pt of l1C.south west] (l1Cbr) {};
 \draw [pattern=north west lines, pattern color=red!20] (l1Ctl) rectangle (l1Cbr);
 \node [anchor = west, below right=60pt and -2pt of l1C.south west] (l1Ctl2) {};
 \node [anchor = west, below right=65pt and 2pt of l1C.south west] (l1Cbr2) {};
 \draw [pattern=north west lines, pattern color=red!20] (l1Ctl2) rectangle (l1Cbr2);
 
 \node [anchor = west, below right=45pt and -2pt of l1D.south west] (l1Dtl) {};
 \node [anchor = west, below right=70pt and 2pt of l1D.south west] (l1Dbr) {};
 \draw [pattern=crosshatch, pattern color=darkgreen!20] (l1Dtl) rectangle (l1Dbr);
 
 \node [anchor = west, below right=30pt and -2pt of l1E.south west] (l1Etl) {};
 \node [anchor = west, below right=60pt and 2pt of l1E.south west] (l1Ebr) {};
 \draw [pattern=north east lines, pattern color=blue!50] (l1Etl) rectangle (l1Ebr);

 \node [anchor=west, right=2.5cm of l1E] (modelNode) {Model $M$};

 \node [anchor = north, below= 2pt of modelNode] (r1) {\scriptsize Pattern for $\Dlp$:};
 \node [anchor = north, below right = -5pt and -10pt of r1.south] (r2) {\scriptsize $\oland(A,\oxor(B,C))$};
 
 \node [anchor = north, below= 10pt of r1] (r3) {\scriptsize Pattern for $\Dlm$:};
 \node [anchor = north, below right = -5pt and -10pt of r3.south] (r4) {\scriptsize $\oland(E)$};
 
 \node [anchor = north, below= 10pt of r3] (r5) {\scriptsize Singletons in $\Data$:};
 \node [anchor = north, below right = -5pt and -10pt of r5.south] (r6) {\scriptsize $F$};

 \end{tikzpicture}
 \caption{\textit{Example database and model.} Left: a toy database $D$ over a set of items, separated by labels into $\Dlp$ and $\Dlm$. Right: the corresponding model $M$ containing patterns describing data partitions $\Dlm$ and $\Dlp$ induced by labels $\lm$ and $\lp$.}\label{fig:toydata}
\end{figure}

Given a dataset of binary transaction data and corresponding binary labels, we aim to find a set of patterns that together identify the partitioning of the data according to the labels.
As an application, consider the sequence of input words of an NLP task as transactions, along with labels that express whether an instance is misclassified by a given model. We are interested in patterns of words that describe these labels to better understand the classifier's errors. In essence, we want to find word combinations such as \OLAND{how, many}, or mutual exclusive patterns, e.g., \OLOR{color, colour}, that capture synonyms or different writing styles, all occuring predominantly when a misclassifcation happens. The pattern language we use is a combination of the two, namely conjunctions of mutual exclusive clauses, e.g., \OLAND{what, \OLOR{color, colour}}. We provide an example in Figure \ref{fig:toydata}.

We thus define a model $M \in \mathcal{M}$ as the set of patterns $\Patterns$ that help to describe given labels.
We look at this problem of identifying a good  model $M$  through the lens of information theory.
In a nutshell, we could consider this problem as transmission of the given data where we assume that the receiver knows the labels of the data. 
We would first send the model containing the patterns, which we then use to send the data to the receiver.
For patterns describing the labels -- the partitions ($\Dlp, \Dlm$) -- well, we can use efficient codes to transmit the database.
If we use too many or overly redundant patterns, or ones that do not describe relevant structures, we spend unnecessary bits in the transmission of the model.
We are, hence, after the model $M^*\in\mathcal{M}$ that minimizes the cost of transmitting data and model, which is captured by the MDL principle.

To ensure that we can always encode any data, $\Model$ contains all singleton words $\Item\in\Items$, describing the entire data $\Data$ without taking labels into account. This model of all singletons also acts as a baseline implementing the assumption that there are no associations that describe the label. 

Let us consider the example in Figure \ref{fig:toydata}, where we would first send $\oland(A,\oxor(B,$ $C))$ occurrences in $\Dlp$, and then its occurrences in $\Dlm$.
Thus, we identify where $A,B,$ and $C$ hold at once, and we leverage the fact that $\oland(A,\oxor(B,C))$ occurs predominantly in $\Dlp$, resulting in more efficient transmission.
Intuitively, a bias of a pattern to occur in one label more than in the other corresponds to a large deviation between the conditional probability -- the pattern occurrence conditioned on the label -- and the unconditional probability -- the pattern occurence in the whole database. 
In this case, the codes are hence more efficient when sending a pattern separately for $\Dlp$ and $\Dlm$.\!\footnote{For a link between probabilities and codelengths, consider for example Shannon Entropy $H(X)=\sum_i P(x_i)\log(P(x_i))$, which gives the minimum number of bits needed to transmit a discrete random variable $X$.}
Coming back to the example, $F$ however occurs similarly often in both data partitions -- there is almost no deviation between conditional and unconditional probability -- hence it is unlikely that it identifies a structural error. Here, the baseline encoding transmitting $F$ as singleton in all of $D$ will be most efficient. 
This approach allows us to identify patterns that occur predominantly for one of the labels as the patterns that yield better compression when conditioned on the labels, and thus characterise labels  in easily understandable terms.

In the following sections, we will formalize this approach using an MDL score to identify that pattern set that best describes the data given the labels. We will first detail how to compute the encoding cost for the data given the model and then the cost for the model itself.

\subsection{Cost of Data Given Model} 
Let us start by explaining how to encode a database \Data\ with singleton items $\Item$ in the absence of any labels, which will later serve as the baseline encoding corresponding to independence between items and labels.
To encode in which transaction an item $\Item$ holds, optimal data-to-model codes are used, which are indices over canonically ordered enumerations~\cite{vitanyi:93:book}. Hence, the data costs are
\begin{align}
\cld\left(\Projection{\Item}{\Data}\mid \Item\right) = \logbinom{|D|}{|\Selection{\Item}{D}|}\;.
\end{align}
Taking into account the partitioning of \Data\ along the label,  yielding \Dlp\ and \Dlm, we encode \Item\ separately:
\begin{align}
\cld\left(\Projection{\Item}{\Data}\mid \Item\right) = \logbinom{|\Dlm|}{|\Selection{\Item}{\Dlm}|} + \logbinom{|\Dlp|}{|\Selection{\Item}{\Dlp}|}\;.
\end{align}
As such, we explicitly reward patterns (here, singletons) that have a different distribution between
the unconditional probability, i.e., its frequency in \Data\,, and the conditional probability of \Item\ conditioned on the label -- i.e., its frequency in \Dlm\ respectively \Dlp. It models the property that we are interested in: patterns that characterize a certain label.
It is straightforward to extend to patterns of co-occurring items $\Pattern = \oland(X_1,\ldots,X_k)$ by selecting on transactions where the pattern holds
\begin{align}
\cld\left(\Projection{\Pattern}{\Data}\mid \Pattern\right) = \logbinom{|\Dlm|}{|\Selection{\Pattern}{\Dlm}|} + \logbinom{|\Dlp|}{|\Selection{\Pattern}{\Dlp}|}\;.
\end{align}
There might be transactions where individual items of \Pattern\ are present, but not all of \Pattern\ holds. To ensure a lossless encoding, the singleton code $\cld(\Projection{\Item}{\Data}\mid \Item)$ is modified to cover all item occurrences left unexplained after transmitting all patterns \Patterns
\[
\cld_s \mkern-2mu \left(\Projection{\Item}{\Data} \mkern-2mu \mid \mkern-2mu \Patterns\right) \mkern-2mu = \mkern-2mu \logbinom{|D|}{|\Selection{\Item}{D} \setminus \big(\bigcup_{\Pattern\in\Patterns, \Item\in\Pattern} ~\Selection{\Pattern}{D}\big)|}\;.
\]
For patterns expressing conjunctions over mutual exclusive items, e.g. $\oland(\oxor(A,B), \oxor(C,D))$,
we first send for both \Dlm\ and \Dlp for which transactions the pattern holds, after which we specify which of the items is active where. We specify items one by one, as we know that when the pattern holds and $A$ is present, $B$ cannot be present too.
With each transmitted item of the clause, there are thus fewer transactions where the remaining items could occur, hence the codelength is reduced.
More formally, the codelength for a pattern $P$ of conjunctions of clauses is given as
\begin{align}
&\cld(\Projection{\Pattern}{\Data} \mid \Pattern) = \sum_{l \in \{-, +\}} \logbinom{|\Dll|}{|\Selection{\Pattern}{\Dll}|} ~+ \\
& \sum_{ \substack{\Clause \in \\ \gamma(\Pattern)}} \sum_{\substack{\Item \in \\  \Clause}} \logbinom{|\Selection{\Pattern}{\Dll}| - \sum_{\Item' \in \Clause, \Item' \leq \Item} |\Selection{\Item'}{\Selection{\Pattern}{\Dll}}|}{|\Selection{\Item}{\Selection{\Pattern}{\Dll}}|}\;,
\end{align}
assuming a canonical order on \Items. With clauses of only length $1$ we arrive at a simple conjunctive pattern, and the function resolves to the codelength function for conjunctive patterns discussed above. Note here that the codelength is the same regardless of the order assumed on the \Items. This statement trivially holds for clauses of length $2$, we provide an argument for the case of $l$ items in the Appendix.

This concludes the definition of codelength functions for transmitting the data. The overall cost of transmitting the data \Data\ given a model $\Model$ is hence
\begin{align*}
 \cl(\Data \mkern-2mu \mid \mkern-2mu \Model) = &\Big( \mkern-4mu \sum_{\Pattern\in\Patterns} \cld(\Projection{\Pattern}{\Data} \mkern-2mu \mid \mkern-2mu \Pattern)\Big)  \mkern-4mu +  \mkern-4mu \Big(\sum_{\Item\in\Items} \cld_s(\Projection{\Item}{\Data}  \mkern-2mu \mid  \mkern-2mu \Patterns)\Big)\;.
\end{align*}

\subsection{Cost of the Model}
Let us now discuss how to transmit the model $\Model$ for pattern set $\Patterns$. First, we transmit the number of patterns $|\Patterns|$ using the MDL-optimal code for integers $\LN(|\Patterns|)$. It is defined as $\LN(n) = \log^*n + \log c_0 $ with $\log^*n = \log n + \log \log n + \ldots$ and $c_0$ being a constant so that $\LN(n)$ satisfies the Kraft-inequality \cite{rissanen:83:integers}.
Then, for each pattern \Pattern, we transmit the number of clauses via $\LN(|\gamma(\Pattern)|)$.
For each such clause, we transmit the items it contains using a log binomial, requiring $\logbinom{|\Items|}{|\Clause|}$ bits plus a parametric complexity term $\LPC(|\Items|)$. The log binomial along with the parametric complexity form the normalized maximum likelihood code for multinomials, which is a refined MDL code. The parametric complexity for multinomials is computable in linear time~\cite{kontkanen:07:linear}.
Lastly, we transmit the parametric complexities of all binomials used in the data encoding.

Combining the above, the overall model cost is 
\begin{align}
	&\cl(\Model) = \; \LN(|\Patterns|) + \sum_{\Pattern\in\Patterns} ( \LN(|\gamma(\Pattern)|) + \LPC(|\Dlp|) +   \\
	&\LPC(|\Dlm|) ) \mkern-2mu + \mkern-7mu \sum_{\Clause\in\Pattern} \mkern-5mu \left( \logbinom{|\Items|}{|\Clause|} \mkern-4mu + \mkern-4mu \LPC(|\Items|)\right) \mkern-4mu +  \mkern-4mu \sum_{\Item\in\Items} \LPC(|\Data|)\;.
\end{align}

\subsection{The Problem, formally}

Based on the above, we can now formally state the problem.

\vspace{0.5em}
\noindent\textsc{Minimal Label Description Problem} \emph{Given data \Data\ over \Items\ and partitions \Dlm\ and \Dlp, find model $\Model\in \Models$ that minimizes the codelength $ \cl(\Model) + \cl(\Data \mid \Model)$.}
\vspace{0.5em}

Solving this problem through enumeration of all models is computationally infeasible as the model space is extremely large (see App. Sec.~\ref{sec:app_modelspace}), and does not lend itself for efficient search.
Hence, we resort to an efficient heuristic for discovering good models.

%% file: approach.tex
\section{ \ourmethod }
\label{sec:method}

To find good pattern sets in practice, we present \ourmethod %
which efficiently explores the search space in a bottom-up heuristic fashion.

\subsection{ Creating and Merging Patterns }
\label{sec:method-merge}
\ourmethod starts with a model \Model that contains only singletons. It then iteratively improves the model by adding, extending, and merging patterns until it can not achieve more gain in the MDL score. To ease the explanation, we first introduce the setting with conjunctive patterns only.
\begin{itemize}[leftmargin=*]
	\setlength\itemsep{-0.2em}
	\item \textit{single items}: $\Item \in \Items$ that improves the MDL score when transmitted separately for \Dlm\ and \Dlp,
	\item \textit{pairs of items}: a new conjunctive pattern $\oland(\Item_1, \Item_2) \in \Items \times \Items$,
	\item \textit{patterns and items}: a new conjunctive pattern $\oland(P, \Item)$ by merging an existing pattern $\Pattern \in \Model$ with an $\Item \in \Items$,
	\item \textit{pairs of patterns}: a new conjunctive pattern $\oland(P_1, P_2)$ obtained by merging two existing patterns $\Pattern_1, \Pattern_2 \in \Model$.  
\end{itemize}
We can speed up the search by pruning infrequent and therewith uninteresting patterns.
Pairs of items for which the transaction sets barely overlap are unlikely to compress well as conjunctive patterns. Hence, we introduce a
minimum overlap threshold of $0.3$ in all experiments.
This straightforwardly leads to algorithm \texttt{createCandidates} that, based on a current model \Model, outputs a set of possible candidate patterns that we will consider as additions to the model. We give pseudocode in the App. Sec.~\ref{app:pseudocode}.

\subsection{Filtering Noise}
\label{sec:fisher}
Additionally to the MDL score, \citet{theory/Fischer20Mexican} proposed to use Fisher's exact test as a filter for spurious patterns.
Here, we use it to test our candidate patterns.
Fisher's exact test allows to assess statistically whether two items co-occur independently based on contingency tables. We assume the hypothesis of homogeneity; in our case that there is no difference in the pattern's probability between \Dlm\ and \Dlp. Fisher showed that the values of the contingency table follow a hypergeometric distribution \cite{theory/Fisher1922Contingency}. We can then compute the p-value for the one-sided test directly via
\begin{align}
p = \sum_{i=0}^{\min(a,d)} \frac{\binom{a+b}{a-i} \binom{c+d}{c+i}}{\binom{n}{a+c}}.
\end{align}
with $c=|\Selection{\Pattern}{\Data}|$, $a=|\Data| - c$, $d=|\Selection{\Pattern}{\Dlp}|$, $b = |\Dlp| - d$ and $n=|\Data|$ for a pattern \Pattern\ labeled with $\lp$. For patterns labeled with $\lm$, the other tail of the distribution is tested (with $a$ and $b$ as well as $c$ and $d$ switching places). In all experiments, we require $p < 0.01$.
A general problem for statistical pattern mining is the lack of an appropriate multiple test correction. 
We, however, only use the test to \textit{filter} candidates, potential false positive patterns passing the test are still evaluated in terms of MDL.

\subsection{The \ourmethod Algorithm}

\begin{algorithm}[h!]
	\caption{\ourmethod} \label{alg:ourmethod}
	\begin{algorithmic}[1]
		\STATE \textbf{Input:} \Data, significance threshold $\alpha$
		\STATE \textbf{Output:} approximation \Model\ of $\Model^*$
		\REPEAT
		\STATE $\Delta' \gets 0$
		\STATE $\Model' \gets \Model$
		\STATE $C \gets $\texttt{createCandidates}$(\Model)$
		\FOR{$\Pattern \in C$}
		\STATE $\Delta \gets \cl(\Data, \Model \oplus \Pattern) - \cl(\Data, \Model)$  \textit{// (neg.) gain}
		\STATE $p \gets $\texttt{FisherExactTest}$(\Pattern)$  \textit{// p-value}
		\IF{$p < \alpha$ and $\Delta < \Delta'$}
		\STATE $\Delta' \gets \Delta$, $\Model' \gets \Model \oplus \Pattern$
		\ENDIF
		\ENDFOR
		\STATE $\Model \gets \Model'$
		\UNTIL{$\Delta' = 0$}
	\end{algorithmic}
\end{algorithm}

Combining the candidate generation and the MDL score from Section \ref{sec:theory}, we obtain \ourmethod. We give the pseudo-code in Algorithm \ref{alg:ourmethod}. Starting with the empty model, we generate candidates and for each of those, we compute the (negative) gain in terms of MDL (line 8) as well as the pattern's p-value (line 9). We select the candidate below a significance threshold $\alpha$ that reaches the highest gain (line 10-12) and add it to the model. If we created the pattern through a merge, we remove its parent patterns from \Model. We repeat the process until no candidate provides further gain in codelength.
As we are after a concise set of patterns that describe the given label, which MDL ensures to find, we analyze \ourmethod's time complexity in terms of the output.
To find a set of $k$ patterns of maximum length $l$ over a dataset of $m$ items, \ourmethod runs in time $O(k\,l\,(k\,l + m^2))$.
For a full derivation and discussion of the algorithm's complexity, see Appendix, Section \ref{sec:app_complexity}.

\subsection{Mutual Exclusivity}

In our practical NLP applications, we are interested, among others, in finding clauses expressing words that are synonyms, that reflect similar concepts, or language variations, such as \OLAND{which,\OLOR{color, colour}} or \OLOR{could, can}.
Such statements, however, require a pattern language beyond purely conjunctive patterns.
We discussed above how to identify the best model over a pattern language of clauses in terms of MDL.
Instead of enumerating all possible clauses exhaustively or searching for an XOR structure like \citet{theory/Fischer20Mexican}, for NLP applications, we follow a more informed approach, taking into account information from pre-trained word embeddings. These classifier-independent embeddings indicate word relations such as similar concepts, synonyms, or writing styles, and hence express exactly the relationships that we are interested in.  In our candidate search, we consider only those XOR terms that span words that are close in the embedding space ($5$ closest neighbours). We give details in App. Sec. ~\ref{sec:app_wordneighbors}.

%% file: experiments.tex
\section{Experiments}

We evaluate our approach on synthetic data with known ground truth, as well as on real world NLP tasks to characterise misclassifications. We compare against significant pattern mining \citep[\textsc{SPuManTe},][]{pellegrina:19:spumante}, rule set mining \citep[\grab,][]{fischer:19:grab}, rule lists \citep[\Classy,][]{proenca2020classy}, top-k subgroup discovery \citep[\Subgroup,][]{subgroup/lemmerich2018pysubgroup}, the subgroup discovery tool \cortana \cite{Meeng2011Cortana,wouter2014scape} and class-specific matrix factorization \citep[\CSalt,][]{hess17CSALT}. As representatives of interpretable, global machine learning models we consider the rule-learner \citep[\ripper,][]{cohen1995fast} and patterns derived from classification trees (\Tree). Due to runtime issues, we compare to the local explainability method \citep[\textsc{Anchors},][]{local/ribeiro18anchor} only in the NER experiment. For similar reason, we exclude \textsc{SliceFinder} \cite{related/Chung2020Slicer}, and disjunctive emerging patterns \cite{emerging/vimieiro2012DisjunctivePattern}; neither completed a single run within 12 hours. Further details are given in Appendix, Section~\ref{sec:app_exp_details}, with datasets and code available online.\!\footnote{\codeurl}

\subsection{Synthetic Data}

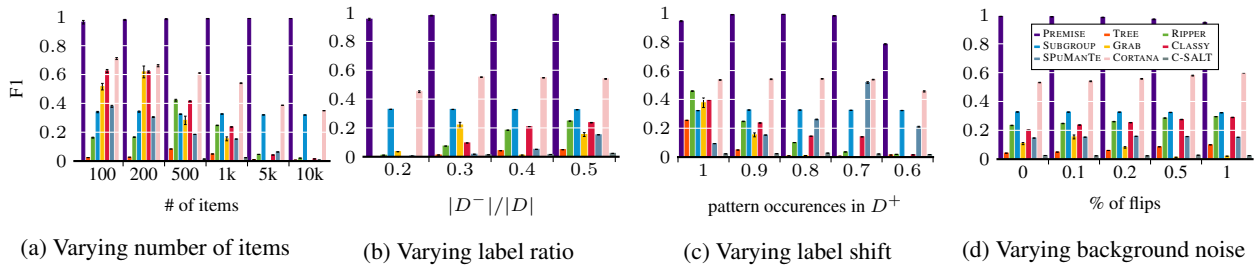
\begin{figure*}
	\centering
	\begin{subfigure}{0.24\textwidth}
		\centering
		\ifpdf
		\begin{tikzpicture}
		\begin{axis}[
		eda ybar,
		xtick=data,
		xticklabels={100,200,500,1k,5k,10k},		
		width = 5cm,
		height = 3.5cm,
		xlabel = {\# of items}, 
		ylabel= {F1},
		extra x ticks={0.5,1.5,...,4.5},
		extra x tick labels={},
		every extra x tick/.style={thin, black, major tick length=2pt},
		bar width = 1.1pt,
		xmin=-0.7, xmax=5.5,
		ymin=0, ymax=1,
		legend columns=2,
		x label style 		= {at={(axis description cs:0.5,-0.05)}, anchor=north, font=\scriptsize},
		y label style 		= {at={(axis description cs:-0.1,0.6)}, rotate=90,  anchor=south, font=\scriptsize},
		legend style={nodes={scale=0.9, transform shape}, at={(0.55,0.92)}, anchor=north, row sep=-1.4pt}
		]
		
		\addplot 
		plot [eda errorbar]
		table[x=num_items, y=mijo, y error=mijo_error] {expres/pure-syn-data_evaluation_gen2-4_num-items_soft-f1_22-05-31_modified-xlabel.txt};
		
		\addplot 
		plot [eda errorbar]
		table[x=num_items, y=tree,y error=tree_error ]{expres/pure-syn-data_evaluation_gen2-4_num-items_soft-f1_22-05-31_modified-xlabel.txt};
		
		\addplot 
		plot [eda errorbar]
		table[x=num_items, y=ripper,y error=ripper_error ]{expres/pure-syn-data_evaluation_gen2-4_num-items_soft-f1_22-05-31_modified-xlabel.txt};
		
		\addplot
		plot [eda errorbar]
		table[x=num_items,y=subgroup, y error=subgroup_error] {expres/pure-syn-data_evaluation_gen2-4_num-items_soft-f1_22-05-31_modified-xlabel.txt};
		
		\addplot
		plot [eda errorbar]
		table[x=num_items,y=realgrab, y error=realgrab_error] {expres/pure-syn-data_evaluation_gen2-4_num-items_soft-f1_22-05-31_modified-xlabel.txt};
		
		\addplot
		plot [eda errorbar]
		table[x=num_items,y=rulelist, y error=rulelist_error] {expres/pure-syn-data_evaluation_gen2-4_num-items_soft-f1_22-05-31_modified-xlabel.txt};
		
		\addplot
		plot [eda errorbar]
		table[x=num_items,y=spumante, y error=spumante_error] {expres/pure-syn-data_evaluation_gen2-4_num-items_soft-f1_22-05-31_modified-xlabel.txt};
		
		\addplot
		plot [eda errorbar]
		table[x=num_items,y=cortana_negr_02quality, y error=cortana_negr_02quality_error] {expres/pure-syn-data_evaluation_gen2-4_num-items_soft-f1_22-05-31_modified-xlabel.txt};
		
		\addplot
		plot [eda errorbar]
		table[x=num_items,y=csalt, y error=csalt_error] {expres/pure-syn-data_evaluation_gen2-4_num-items_soft-f1_22-05-31_modified-xlabel.txt};
		
		\end{axis}
		\end{tikzpicture}
		\fi
		\caption{Varying number of items\label{fig:pure-syn-data_num-items}}
	\end{subfigure}
	\begin{subfigure}{0.24\textwidth}
		\centering
		\ifpdf
		\begin{tikzpicture}
		\begin{axis}[
		eda ybar,
		xtick=data,
		width = 5cm,
		height = 3.5cm,
		xlabel = {$|\Dlm| / |\Data|$},
		ylabel={},
		extra x ticks={0.25,0.35,...,0.55},
		extra x tick labels={},
		every extra x tick/.style={thin, black, major tick length=2pt},
		enlarge x limits=0.19,
		ymin=0, ymax=1,
		legend columns=2,
		x label style 		= {at={(axis description cs:0.5,-0.05)}, anchor=north, font=\scriptsize},
		y label style 		= {at={(axis description cs:-0.1,0.6)}, rotate=90,  anchor=south, font=\scriptsize},
		legend style={nodes={scale=0.8, transform shape}, at={(0.5,0.95)}, anchor=north, row sep=-1.6pt}
		]
		
		\addplot 
		plot [eda errorbar]
		table[x=l1_percentage, y=mijo, y error=mijo_error] {expres/pure-syn-data_evaluation_gen2-1_l1-percentage_soft-f1_22-05-31.txt};
		
		\addplot 
		plot [eda errorbar]
		table[x=l1_percentage, y=tree,y error=tree_error ]{expres/pure-syn-data_evaluation_gen2-1_l1-percentage_soft-f1_22-05-31.txt};
		
		\addplot 
		plot [eda errorbar]
		table[x=l1_percentage, y=ripper,y error=ripper_error ]{expres/pure-syn-data_evaluation_gen2-1_l1-percentage_soft-f1_22-05-31.txt};
		
		\addplot
		plot [eda errorbar]
		table[x=l1_percentage,y=subgroup, y error=subgroup_error] {expres/pure-syn-data_evaluation_gen2-1_l1-percentage_soft-f1_22-05-31.txt};
		
		\addplot
		plot [eda errorbar]
		table[x=l1_percentage,y=realgrab, y error=realgrab_error] {expres/pure-syn-data_evaluation_gen2-1_l1-percentage_soft-f1_22-05-31.txt};
		
		\addplot
		plot [eda errorbar]
		table[x=l1_percentage,y=rulelist, y error=rulelist_error] {expres/pure-syn-data_evaluation_gen2-1_l1-percentage_soft-f1_22-05-31.txt};
		
		\addplot
		plot [eda errorbar]
		table[x=l1_percentage,y=spumante, y error=spumante_error] {expres/pure-syn-data_evaluation_gen2-1_l1-percentage_soft-f1_22-05-31.txt};
		
		\addplot
		plot [eda errorbar]
		table[x=l1_percentage,y=cortana_negr_02quality, y error=cortana_negr_02quality_error] {expres/pure-syn-data_evaluation_gen2-1_l1-percentage_soft-f1_22-05-31.txt};
		
		\addplot
		plot [eda errorbar]
		table[x=l1_percentage,y=csalt, y error=csalt_error] {expres/pure-syn-data_evaluation_gen2-1_l1-percentage_soft-f1_22-05-31.txt};

		\end{axis}
		\end{tikzpicture}
		\fi
		\caption{Varying label ratio \label{fig:pure-syn-data_l1-percentage}}
	\end{subfigure}
	\begin{subfigure}{0.24\textwidth}
		\centering
		\ifpdf
		\begin{tikzpicture}
		\begin{axis}[
		eda ybar,
		xtick=data,
		x dir=reverse,
		width = 5cm,
		height = 3.5cm,
		xlabel = {pattern occurences in $\Dlp$}, 
		ylabel= {},
		extra x ticks={1.05,0.95,...,0.55},
		extra x tick labels={},
		every extra x tick/.style={thin, black, major tick length=2pt},
		enlarge x limits=0.13,
		bar width = 1.4pt,
		ymin=0, ymax=1,
		legend columns=2,
		x label style 		= {at={(axis description cs:0.5,-0.05)}, anchor=north, font=\scriptsize},
		y label style 		= {at={(axis description cs:-0.1,0.5)}, anchor=south, font=\scriptsize},
		legend style={nodes={scale=0.8, transform shape}, at={(0.35,0.9)}, anchor=north, row sep=-1.4pt}
		]
		
		\addplot 
		plot [eda errorbar]
		table[x=shift_value, y=mijo, y error=mijo_error] {expres/pure-syn-data_evaluation_gen2-2_label-shift_soft-f1_22-05-31.txt};

		\addplot 
		plot [eda errorbar]
		table[x=shift_value, y=tree,y error=tree_error ]{expres/pure-syn-data_evaluation_gen2-2_label-shift_soft-f1_22-05-31.txt};
		
		\addplot 
		plot [eda errorbar]
		table[x=shift_value, y=ripper,y error=ripper_error ]{expres/pure-syn-data_evaluation_gen2-2_label-shift_soft-f1_22-05-31.txt};
		
		\addplot
		plot [eda errorbar]
		table[x=shift_value,y=subgroup, y error=subgroup_error] {expres/pure-syn-data_evaluation_gen2-2_label-shift_soft-f1_22-05-31.txt};
		
		\addplot
		plot [eda errorbar]
		table[x=shift_value,y=realgrab, y error=realgrab_error] {expres/pure-syn-data_evaluation_gen2-2_label-shift_soft-f1_22-05-31.txt};
		
		\addplot
		plot [eda errorbar]
		table[x=shift_value,y=rulelist, y error=rulelist_error] {expres/pure-syn-data_evaluation_gen2-2_label-shift_soft-f1_22-05-31.txt};
		
		\addplot
		plot [eda errorbar]
		table[x=shift_value,y=spumante, y error=spumante_error] {expres/pure-syn-data_evaluation_gen2-2_label-shift_soft-f1_22-05-31.txt};
		
		\addplot
		plot [eda errorbar]
		table[x=shift_value,y=cortana_negr_02quality, y error=cortana_negr_02quality_error] {expres/pure-syn-data_evaluation_gen2-2_label-shift_soft-f1_22-05-31.txt};
		
		\addplot
		plot [eda errorbar]
		table[x=shift_value,y=csalt, y error=csalt_error] {expres/pure-syn-data_evaluation_gen2-2_label-shift_soft-f1_22-05-31.txt};
		
		\end{axis}
		\end{tikzpicture}
		\fi
		\caption{Varying label shift\label{fig:pure-syn-data_shift}}
	\end{subfigure}
	\begin{subfigure}{0.24\textwidth}
		\centering
		\ifpdf
		\begin{tikzpicture}
		\begin{axis}[
		eda ybar,
		xtick=data,
		xticklabels={0,0.1,0.2,0.5,1},		
		width = 5cm,
		height = 3.5cm,
		xlabel = {\% of flips}, 
		ylabel= {},
		extra x ticks={0.5,1.5,...,3.5},
		extra x tick labels={},
		every extra x tick/.style={thin, black, major tick length=2pt},
		enlarge x limits=0.13,
		bar width = 1.4pt,
		ymin=0, ymax=1,
		legend columns=2,
		x label style 		= {at={(axis description cs:0.5,-0.05)}, anchor=north, font=\scriptsize},
		y label style 		= {at={(axis description cs:-0.1,0.5)}, anchor=south, font=\scriptsize},
		legend columns=3,
		legend style={nodes={scale=0.60, transform shape}, at={(0.5,0.94)}, anchor=north, row sep=-1.6pt}
		]
		
		\addplot 
		plot [eda errorbar]
		table[x=noise_rate, y=mijo, y error=mijo_error] {expres/pure-syn-data_evaluation_gen2-3_background-noise-rate_soft-f1_22-05-31_modified-xlabel.txt};
		
		\addplot 
		plot [eda errorbar]
		table[x=noise_rate, y=tree,y error=tree_error ]{expres/pure-syn-data_evaluation_gen2-3_background-noise-rate_soft-f1_22-05-31_modified-xlabel.txt};
		
		\addplot 
		plot [eda errorbar]
		table[x=noise_rate, y=ripper,y error=ripper_error ]{expres/pure-syn-data_evaluation_gen2-3_background-noise-rate_soft-f1_22-05-31_modified-xlabel.txt};	
		
		\addplot
		plot [eda errorbar]
		table[x=noise_rate,y=subgroup, y error=subgroup_error] {expres/pure-syn-data_evaluation_gen2-3_background-noise-rate_soft-f1_22-05-31_modified-xlabel.txt};
		
		\addplot
		plot [eda errorbar]
		table[x=noise_rate,y=realgrab, y error=realgrab_error] {expres/pure-syn-data_evaluation_gen2-3_background-noise-rate_soft-f1_22-05-31_modified-xlabel.txt};
		
		\addplot
		plot [eda errorbar]
		table[x=noise_rate,y=rulelist, y error=rulelist_error] {expres/pure-syn-data_evaluation_gen2-3_background-noise-rate_soft-f1_22-05-31_modified-xlabel.txt};
		
		\addplot
		plot [eda errorbar]
		table[x=noise_rate,y=spumante, y error=spumante_error] {expres/pure-syn-data_evaluation_gen2-3_background-noise-rate_soft-f1_22-05-31_modified-xlabel.txt};
		
		\addplot
		plot [eda errorbar]
		table[x=noise_rate,y=cortana_negr_02quality, y error=cortana_negr_02quality_error] {expres/pure-syn-data_evaluation_gen2-3_background-noise-rate_soft-f1_22-05-31_modified-xlabel.txt};
		
		\addplot
		plot [eda errorbar]
		table[x=noise_rate,y=csalt, y error=csalt_error] {expres/pure-syn-data_evaluation_gen2-3_background-noise-rate_soft-f1_22-05-31_modified-xlabel.txt};
		
		\legend{\ourmethod, \Tree, \ripper, \textsc{Subgroup}, \Grab, \Classy, \Spumante, \cortana, \CSalt}
		
		\end{axis}
		\end{tikzpicture}
		\fi
		\caption{Varying background noise\label{fig:pure-syn-data_background-noise}}
	\end{subfigure}
	\caption{\textit{Synthetic data results.} As competitors only recover fragments of patterns, the results are in terms of a soft F1 score, which also rewards the discovery of fragments, as defined in Appendix, Section \ref{sec:app_f1}.\label{fig:pure-syn-data}}.
	\vspace{-0.5cm}
\end{figure*}

\begin{figure*}[h]
	\centering
	\begin{subfigure}{0.32\textwidth}
		\ifpdf
		\begin{tikzpicture}
		\begin{axis}[
		eda ybar,
		xtick=data,
		width = 6.3cm,
		height = 3.5cm,
		xlabel = {$|\Pattern|$}, 
		ylabel= {F1},
		extra x ticks={1.5,...,7.5},
		extra x tick labels={},
		every extra x tick/.style={thin, black, major tick length=2pt},
		enlarge x limits=0.1,
		bar width = 1.1pt,
		ymin=0, ymax=1,
		legend columns=2,
		x label style 		= {at={(axis description cs:0.5,-0.05)}, anchor=north, font=\scriptsize},
		y label style 		= {at={(axis description cs:-0.1,0.5)}, anchor=south, font=\scriptsize},
		legend style={nodes={scale=0.8, transform shape}, at={(0.4,0.75)}, anchor=north, row sep=-1.4pt}
		]
		
		\addplot 
		plot [eda errorbar]
		table[x=patternlength,y=ptb-dev-onlyand3, y error=ptb-dev-onlyand3_error] {expres/syn-data_evaluation_ptb-dev-onlyand_clause-lengths_no-noise_f1_22-05-31.txt};
		
		\addplot 
		plot [eda errorbar]
		table[x=patternlength, y=ptb-dev-onlyand3_tree,y error=ptb-dev-onlyand3_tree_error ]{expres/syn-data_evaluation_ptb-dev-onlyand_clause-lengths_no-noise_f1_22-05-31.txt};
		
		\addplot 
		plot [eda errorbar]
		table[x=patternlength, y=ptb-dev-onlyand3_ripper,y error=ptb-dev-onlyand3_ripper_error ]{expres/syn-data_evaluation_ptb-dev-onlyand_clause-lengths_no-noise_f1_22-05-31.txt};
		
		\addplot
		plot [eda errorbar]
		table[x=patternlength,y=ptb-dev-onlyand3_subgroup, y error=ptb-dev-onlyand3_subgroup_error] {expres/syn-data_evaluation_ptb-dev-onlyand_clause-lengths_no-noise_f1_22-05-31.txt};
		
		\addplot
		plot [eda errorbar]
		table[x=patternlength,y=ptb-dev-onlyand3_realgrab, y error=ptb-dev-onlyand3_realgrab_error] {expres/syn-data_evaluation_ptb-dev-onlyand_clause-lengths_no-noise_f1_22-05-31.txt};
		
		\addplot
		plot [eda errorbar]
		table[x=patternlength,y=ptb-dev-onlyand3_rulelist, y error=ptb-dev-onlyand3_rulelist_error] {expres/syn-data_evaluation_ptb-dev-onlyand_clause-lengths_no-noise_f1_22-05-31.txt};
		
		\addplot
		plot [eda errorbar]
		table[x=patternlength,y=ptb-dev-onlyand3_spumante, y error=ptb-dev-onlyand3_spumante_error] {expres/syn-data_evaluation_ptb-dev-onlyand_clause-lengths_no-noise_f1_22-05-31.txt};
		
		\addplot
		plot [eda errorbar]
		table[x=patternlength,y=ptb-dev-onlyand3_cortana_negr, y error=ptb-dev-onlyand3_cortana_negr_error] {expres/syn-data_evaluation_ptb-dev-onlyand_clause-lengths_no-noise_f1_22-05-31.txt};
		
		\addplot
		plot [eda errorbar]
		table[x=patternlength,y=ptb-dev-onlyand3_csalt-step10, y error=ptb-dev-onlyand3_csalt-step10_error] {expres/syn-data_evaluation_ptb-dev-onlyand_clause-lengths_no-noise_f1_22-05-31.txt};
		
		\end{axis}
		\end{tikzpicture}
		\fi
		\caption{Varying pattern length\label{fig:syn-data_onlyand_pattern-length}}
	\end{subfigure}
	\begin{subfigure}{0.2\textwidth}
		\ifpdf
		\begin{tikzpicture}
		\begin{axis}[
		eda ybar,
		xtick=data,
		x dir=reverse,
		width = 4.5cm,
		height = 3.5cm,
		xlabel = {Noise: shift}, 
		ylabel= {},
		extra x ticks={1.05,0.95,...,0.55},
		extra x tick labels={},
		every extra x tick/.style={thin, black, major tick length=2pt},
		enlarge x limits=0.13,
		ymin=0, ymax=1,
		bar width = 1.1pt,
		legend columns=2,
		x label style 		= {at={(axis description cs:0.5,-0.05)}, anchor=north, font=\scriptsize},
		y label style 		= {at={(axis description cs:-0.1,0.5)}, anchor=south, font=\scriptsize},
		legend style={nodes={scale=0.6, transform shape}, at={(0.55,0.8)}, anchor=north, row sep=-1.4pt}
		]
		
		\addplot 
		plot [eda errorbar]
		table[x=patternlength,y=ptb-dev-onlyand1, y error=ptb-dev-onlyand1_error] {expres/syn-data_evaluation_ptb-dev-onlyand_bias-shift_f1_22-05-31.txt};
		
		\addplot 
		plot [eda errorbar]
		table[x=patternlength, y=ptb-dev-onlyand1_tree,y error=ptb-dev-onlyand1_tree_error ] {expres/syn-data_evaluation_ptb-dev-onlyand_bias-shift_f1_22-05-31.txt};
		
		\addplot 
		plot [eda errorbar]
		table[x=patternlength, y=ptb-dev-onlyand1_ripper,y error=ptb-dev-onlyand1_ripper_error ] {expres/syn-data_evaluation_ptb-dev-onlyand_bias-shift_f1_22-05-31.txt};
		
		\addplot
		plot [eda errorbar]
		table[x=patternlength,y=ptb-dev-onlyand1_subgroup, y error=ptb-dev-onlyand1_subgroup_error] {expres/syn-data_evaluation_ptb-dev-onlyand_bias-shift_f1_22-05-31.txt};
		
		\addplot
		plot [eda errorbar]
		table[x=patternlength,y=ptb-dev-onlyand1_realgrab, y error=ptb-dev-onlyand1_realgrab_error] {expres/syn-data_evaluation_ptb-dev-onlyand_bias-shift_f1_22-05-31.txt};
		
		\addplot
		plot [eda errorbar]
		table[x=patternlength,y=ptb-dev-onlyand1_rulelist, y error=ptb-dev-onlyand1_rulelist_error] {expres/syn-data_evaluation_ptb-dev-onlyand_bias-shift_f1_22-05-31.txt};
		
		\addplot
		plot [eda errorbar]
		table[x=patternlength,y=ptb-dev-onlyand1_spumante, y error=ptb-dev-onlyand1_spumante_error] {expres/syn-data_evaluation_ptb-dev-onlyand_bias-shift_f1_22-05-31.txt};
		
		\addplot
		plot [eda errorbar]
		table[x=patternlength,y=ptb-dev-onlyand1_cortana_negr, y error=ptb-dev-onlyand1_cortana_negr_error] {expres/syn-data_evaluation_ptb-dev-onlyand_bias-shift_f1_22-05-31.txt};
		
		\addplot
		plot [eda errorbar]
		table[x=patternlength,y=ptb-dev-onlyand1_csalt-step10_error, y error=ptb-dev-onlyand1_csalt-step10_error] {expres/syn-data_evaluation_ptb-dev-onlyand_bias-shift_f1_22-05-31.txt};
		
		\legend{\ourmethod, \Tree, \ripper, \textsc{Subgroup}, \Grab, \Classy, \Spumante, \cortana, \CSalt}
		\end{axis}
		\end{tikzpicture}
		\fi
		\caption{Varying shift noise\label{fig:syn-data_onlyand_shift}}
	\end{subfigure}
	\begin{subfigure}{0.2\textwidth}
		\ifpdf
		\begin{tikzpicture}
		\begin{axis}[
		eda ybar,
		xtick=data,
		width = 4.5cm,
		height = 3.5cm,
		xlabel = {Noise: miscl. without pattern}, 
		ylabel= {},
		extra x ticks={0.025,0.075,...,0.2},
		extra x tick labels={},
		every extra x tick/.style={thin, black, major tick length=2pt},
		enlarge x limits=0.13,
		bar width = 1.1pt,
		ymin=0, ymax=1,
		legend columns=2,
		x label style 		= {at={(axis description cs:0.5,-0.05)}, anchor=north, font=\scriptsize},
		y label style 		= {at={(axis description cs:-0.1,0.5)}, anchor=south, font=\scriptsize},
		legend style={nodes={scale=0.9, transform shape}, at={(0.75,0.3)}, anchor=north, row sep=-1.4pt}
		]
		
		\addplot 
		plot [eda errorbar]
		table[x=patternlength,y=ptb-dev-onlyand1, y error=ptb-dev-onlyand1_error] {expres/syn-data_evaluation_ptb-dev-onlyand_misclass-without-pattern_f1_22-05-31.txt};
		
		\addplot 
		plot [eda errorbar]
		table[x=patternlength, y=ptb-dev-onlyand1_tree,y error=ptb-dev-onlyand1_tree_error ] {expres/syn-data_evaluation_ptb-dev-onlyand_misclass-without-pattern_f1_22-05-31.txt};
		
		\addplot 
		plot [eda errorbar]
		table[x=patternlength, y=ptb-dev-onlyand1_ripper,y error=ptb-dev-onlyand1_ripper_error ] {expres/syn-data_evaluation_ptb-dev-onlyand_misclass-without-pattern_f1_22-05-31.txt};
		
		\addplot
		plot [eda errorbar]
		table[x=patternlength,y=ptb-dev-onlyand1_subgroup, y error=ptb-dev-onlyand1_subgroup_error] {expres/syn-data_evaluation_ptb-dev-onlyand_misclass-without-pattern_f1_22-05-31.txt};
		
		\addplot
		plot [eda errorbar]
		table[x=patternlength,y=ptb-dev-onlyand1_realgrab, y error=ptb-dev-onlyand1_realgrab_error] {expres/syn-data_evaluation_ptb-dev-onlyand_misclass-without-pattern_f1_22-05-31.txt};
		
		\addplot
		plot [eda errorbar]
		table[x=patternlength,y=ptb-dev-onlyand1_rulelist, y error=ptb-dev-onlyand1_rulelist_error] {expres/syn-data_evaluation_ptb-dev-onlyand_misclass-without-pattern_f1_22-05-31.txt};
		
		\addplot
		plot [eda errorbar]
		table[x=patternlength,y=ptb-dev-onlyand1_spumante, y error=ptb-dev-onlyand1_spumante_error] {expres/syn-data_evaluation_ptb-dev-onlyand_misclass-without-pattern_f1_22-05-31.txt};
		
		\addplot
		plot [eda errorbar]
		table[x=patternlength,y=ptb-dev-onlyand1_cortana_negr, y error=ptb-dev-onlyand1_cortana_negr_error] {expres/syn-data_evaluation_ptb-dev-onlyand_misclass-without-pattern_f1_22-05-31.txt};
		
		\addplot
		plot [eda errorbar]
		table[x=patternlength,y=ptb-dev-onlyand1_csalt-step10, y error=ptb-dev-onlyand1_csalt-step10_error] {expres/syn-data_evaluation_ptb-dev-onlyand_misclass-without-pattern_f1_22-05-31.txt};
		
		\end{axis}
		\end{tikzpicture}
		\fi
		\caption{Varying label noise\label{fig:syn-data_onlyand_misclass-without-pattern}}
	\end{subfigure}
	\begin{subfigure}{0.24\textwidth}
		\ifpdf
		\begin{tikzpicture}
		\begin{axis}[
		eda line,
		scaled x ticks=false,
		xtick=data,
		width = 5cm,
		height = 3.5cm,
		xlabel = {Noise: miscl. without pattern}, 
		ylabel= {},
		ymin=0, ymax=1,
		x label style 		= {at={(axis description cs:0.5,-0.15)}, anchor=north, font=\scriptsize},
		y label style 		= {at={(axis description cs:-0.1,0.5)}, anchor=south, font=\scriptsize},
		legend style={nodes={scale=0.9, transform shape}, at={(0.75,0.6)}, anchor=north, row sep=-1.4pt}
		]
		
		\addplot 
		plot [eda errorline]
		table[x=noise,y=shift1, y error=shift1_error] {expres/syn-data_evaluation_and+or-data_mijo_f1_21-02-08.txt};
		
		\addplot 
		plot [eda errorline]
		table[x=noise, y=shift08,y error=shift08_error ]{expres/syn-data_evaluation_and+or-data_mijo_f1_21-02-08.txt};
		
		\addplot
		plot [eda errorline]
		table[x=noise,y=shift06, y error=shift06_error] {expres/syn-data_evaluation_and+or-data_mijo_f1_21-02-08.txt};
		
		\legend{Noise: shift 1, Noise: shift 0.8, Noise: shift 0.6}
		\end{axis}
		\end{tikzpicture}
		\fi
		\caption{Complex clauses\label{fig:syn-data_and+or}}
	\end{subfigure}
	\caption{\textit{Synthetic text data results.} On synthetic text data, varying the number of items per pattern (a), the amount of \textit{shift noise} (b), and the amount of \textit{label noise} (c), we visualize the results in terms of F1 score with respect to the ground truth for existing methods and \ourmethod. We additionally provide results of \ourmethod on data containing patterns of mutual exclusive clauses for varying \textit{shift noise} (d).}\label{fig:syn-text-data-results}
\end{figure*}
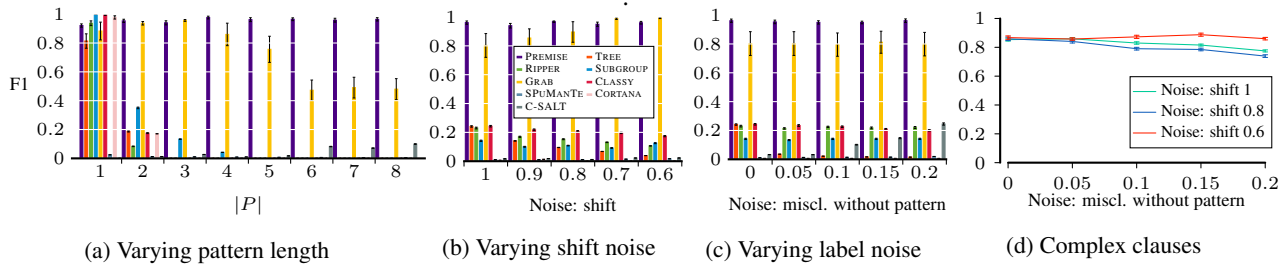

To evaluate against a known ground truth, we generate synthetic data where we insert patterns. Unless specified differently, for each of the experiments we generate a data matrix with $10\,000$ samples, half of which get label $\lm$. The set of items \Items has size $1000$. We draw patterns of length 2-5 from \Items with replacement until $50\%$ of items are covered.
For each pattern we then draw $k\sim\mathcal{N}(150,20)$ and set the items of the pattern in $.9k$ random transactions from $\Dlp$, and $.1k$ transaction from $\Dlm$ to $1$. This corresponds to a typical sparsity level for pattern mining problems.
Additionally, for each item that is part in a pattern, we let it occur in $k\sim\mathcal{N}(50,20)$ random transactions from $\Data$.
For all items not part of a pattern, we let them occur in $k\sim\mathcal{N}(150,20)$ transactions from $\Data$.
Lastly, we introduce background noise by flipping $.1\%$ of the matrix values.

We evaluate all methods with respect to \emph{scalability} (size of item sets \Items), \emph{label imbalance} (proportion of transactions having label $\lm$), \emph{label shift} (patterns occurring not exclusively in one of the labels) and \emph{robustness to background noise} (flipping a fraction of entries of the data matrix). The results are shown in Figure \ref{fig:pure-syn-data}. 
The performance of most existing methods deteriorates already for data with several hundred items. We observe similar effects for increasing label imbalance which is e.g. encountered in misclassified samples that make up only a small fraction of overall samples. For label shift, we adapt the occurrence of patterns between $1$, meaning the pattern occurs exclusively in one partition of the database, to $.6$, meaning that only 60\% of the transaction where a pattern occurs have one label. Again, most baselines struggle with this setting. \CSalt yields relatively good precision for balanced data but its recall is close to 0. \Subgroup, \cortana and \ourmethod are robust to these changes. In most cases, however, \Subgroup and \cortana yield (soft) F1 scores of less than $.4$ and $.6$, respectively, while \ourmethod gives scores close to $1$.

\smallskip\noindent\emph{Synthetic text data.}
For an evaluation with known ground truth more similar to the NLP application domain, we evaluate how well all methods cope with item -- or word -- distributions similar to real text. We report these experiments in Fig.~\ref{fig:syn-text-data-results} with more details in \ref{sec:app_syn_text}. For most most baselines performance quickly deteriorates for longer patterns. \Grab is able to retrieve longer patterns and is resistant to shift and noise in the form of non-systematic label errors. \ourmethod outperforms all competitors, achieving consistently high F1 scores beyond $.92$. For complex patterns consisting of conjunctive clauses of disjunctions, we verify that \ourmethod is able to retrieve them even in the presence of noise.

\subsection{Real Data: VQA}
\label{sec:experiments_vqa}

Visual Question Answering (VQA) is the popular and challenging task of answering textual questions about a given image. We analyze the misclassification of Visual7W \cite{vqa/Zhu2016Visual7W} and the state-of-the-art LXMERT \cite{vqa/Tan2019LXMERT}, both specific architectures for different VQA tasks. Visual7W reaches 54\% accuracy in 4-option multiple choice, LXMERT a validation score of 70\%. Both classifiers perform far from optimal and thus serve as interesting applications for describing (misclassification) labels. We derive misclassification data sets from applying the classifiers to the development sets.

\Spumante as well as \Tree retrieve several hundred or thousand patterns making it difficult to interpret the results.
Furthermore, we know from the previous experiments that these methods find thousands of patterns even when there exist only few ground truth patterns.
\Subgroup requires the user to specify the number of patterns a-priori, which is not known. The discovered patterns are highly redundant with often ten or more patterns expressing the same cause for misclassification.
It is thus hard to get a full description of what goes wrong, it lacks the power of set mining approaches that evaluate patterns \textit{together}. \cortana filters more strongly, the discovered patterns are, however, still redundant. \CSalt finds a few, partially redundant patterns. Most patterns found by \Classy consist of only one token, \grab\ and \ripper fail to retrieve meaningful results. %
In Appendix, Table~\ref{tab:real-data_stats}, we provide statistics about the data and retrieved patterns, a full list of retrieved patterns for all methods can be found in the online material.

The patterns found by \ourmethod (see Tab.~\ref{tab:examples_vqa}, full results online), highlight the advantage of the richer pattern language, allowing to find patterns with related concepts such as \OLAND{what, \OLOR{color, colors, colour}}.
Generally, the patterns found by \ourmethod highlight different types of wrongly answered questions, including counting questions, identification of objects and their colors, spatial reasoning, and higher reasoning tasks like reading signs.
Furthermore, \ourmethod retrieves both frequent patterns, such as \OLAND{how, many} and %
rare patterns such as \OLAND{on, wall, hanging}.  %

\ourmethod also discovers patterns that are biased towards correct classification, which can indicate issues with the dataset. For instance, \OLAND{who, took, \OLOR{photo, picture, pic, photos, photograph}}, although a difficult question, is nearly always answered by ''photographer`` and thus easy to learn.
Another problematic question is indicated by the pattern \OLAND{clock, time}, where usually the answer is ''UNK`` -- the unknown word token -- due to the limited vocabulary of Visual7W.
The pattern hence indicates a setting where the VQA classifier undeservedly gets a good score.

By adding additional information as items to each instance, it is possible to gain further insights.
Appending the correct output to each instance, we observe for the question when the picture was taken two different trends. On the one hand, the discovered pattern \OLAND{when, \OLOR{daytime, nighttime}} is associated with correct classification, the pattern \OLAND{when, \OLOR{evening, morning, afternoon, lunchtime}}, on the other hand, points towards misclassification. This is intuitively consistent as the answers ``daytime'' and ``nighttime'' are easier to choose based on a picture.

We observe in the discovered patterns that the Visual7W and LXMERT classifiers share certain issues, like the counting questions. However, no patterns regarding color or spatial position are retrieved. This seems to indicate that the more recent LXMERT classifier can handle these better. %

\begin{table}
	\caption{\textit{VQA example patterns for Visual7W found by \ourmethod.} %
	More examples in Table \ref{tab:app_lxmert_examples_vqa} in the Appendix. \label{tab:examples_vqa}}
	\small
	\centering
			\vspace{0pt}
			\begin{tabular}{p{3.5cm}|l}
				\toprule
				pattern & example from the dataset\\ \midrule
				\textit{UNK} & how are the UNK covered \\
				\OLAND{how, many} & how many elephants are there \\
				$\oland$(\textit{what, }$\oxor$(\textit{color, } & what color is the bench \\
				\hspace{0.31cm} \textit{colors, colour})) & \\
				\OLAND{on, top, of} & what is on the top of the cake\\
				\OLAND{left, to} & what can be seen to the left\\
				\OLAND{on, wall, hanging} & what is hanging on the wall\\
				\OLAND{how, does, look} & how does the woman look\\
				$\oland$(\textit{what, does, }$\oxor$(\textit{say, } & what does the sign say\\
				\textit{like, think, know, want})) & \\
				\bottomrule
			\end{tabular}
\end{table}

\subsection{Real Data: NER}

A machine learning classifier might perform well during development, its performance when deployed ``in the wild'' however is often much worse. Understanding the difference is important for being able to improve the classifier. 
Here, we investigate the popular LSTM+CNN+CRF architecture \cite{ner/ma2016lstmcnncrf} for Named Entity Recognition (NER). 
The classifier is trained on the standard NER dataset CoNLL03, where it achieves a good performance (F1-score of 0.93). On OntoNotes, a dataset covering a wider range of topics,
the performance drops to 0.61 F1 on the development set. We evaluate on this split of the data consisting of $16$k sentences and $23$k unique items.

\Anchors allows to obtain conjunctive patterns to explain NLP instances locally. It took, however, several days to analyze all misclassifications on modern GPU hardware due to the necessary, repeated queries to the NER classifier. \Anchors finds 4.1k unique patterns, many of which are redundant, overly long, and specific. As expected from a local method, the patterns are highly specific and thus identify problems of the model for particular instances rather than identifying the the general issues.
\ourmethod retrieves a concise set of 190 patterns. An example is \OLAND{\mbox{-LRB-}, \mbox{-RRB-}} that indicates different preprocessing of the text. %
Patterns also indicate problems with differing labeling conventions, e.g. the patterns \OLAND{'s}  and \OLAND{Wall, Street}, which are handled differently for entities in OntoNotes. %
We can also isolate issues with OntoNotes alone, which contains bible excerpts that are not labeled at all. We discover this through patterns that describe this domain (\textit{God}, \textit{Jesus}, \textit{Samuel}).

To empirically validate that the found patterns affect the classifier's performance, we select the top 50 patterns according to gain in MDL and for each pattern sample 5 sentences containing it uniformly at random from the OntoNotes training data. The CoNLL03 classifier is then fine-tuned on this data. Sampling and fine-tuning is repeated $20$ times with different seeds. Using the pattern-guided data, the performance is improved to $0.67$ mean F1 score (SE $0.003$) compared to sampling fully at random where only a small improvement to $0.62$ (SE $0.005$) is achieved. This shows that the patterns discovered by \ourmethod provide actionable insights into how a classifier can be improved.

%% file: conclusion.tex
\section{Discussion}

Experiments show that \ourmethod provides concise and interpretable descriptions of labeled data.
On synthetic data, we find that the state-of-the-art methods across different fields have severe difficulties finding the ground truth pattern set that describe the given labels. \ourmethod is the only approach that is at the same time robust to noise, label imbalance, and easily scaling to thousands of items.
When considering as label the information on whether a sample has been misclassified or not by a particular method, these approaches allow to find descriptions of which patterns in the data are correlated with misclassifications.
For tasks like characterising misclassifications of NLP models, however, the labels are inherently imbalanced and the sets of items -- in this case tokens -- is large.
To capture the structures of word associations, we further need a richer pattern language capturing mutual exclusiveness, which only \ourmethod provides.

On two models for VQA, we set for characterising their misclassifications. While some of the competing methods retrieve reasonable explanations, these are highly redundant %
with several hundred or thousand patterns.
Moreover, important concepts, such as patterns that are similar across related words or synonyms, are completely missed.
\ourmethod, on the other hand, discovers succinct sets of patterns that provide interesting characterizations, revealing that models struggle with counting, spatial orientation, reading, and identifies shortcomings in training data. For a popular NER classifier, we consider a model applied to text of a different source and characterize the resulting classification errors. Compared to the local explanation method \Anchors, \ourmethod retrieves a more succinct set of patterns in less time and we show that the obtained insights are actionable.

To show that the pattern sets are not only interpretable and give interesting insights, but also actionable, we analyze a popular classifier for Named Entity Recognition.
In particular, we consider a model applied to text of a different source and characterize the resulting classification errors with \ourmethod, and compare it with the recent local explanation method \Anchors.
While \ourmethod is able to retrieve a pattern set swiftly in few hours on commodity hardware, \Anchors requires several days on a modern GPU to deliver results.
Inspecting the retrieved patterns confirms that also for NER models \ourmethod is able to retrieve meaningful patterns explaining misclassification, while \Anchors finds a large set of overly long and redundant patterns. Furthermore, as expected from a local method, the patterns describe instance-specific problems of the model rather than identifying the general issues that the model has.

\section{Conclusion}

We considered the problem of finding interpretable and non-redundant descriptions of a given label, and proposed to discover pattern sets to describe the labels based on the Minimum Description Length Principle. To solve this formulation in practice, we suggested an efficient bottom-up heuristic \ourmethod.
Our method showed to be the only approach that scales well to data typical in real world problem settings, while at the same time being robust to noise, and label imbalance.
With these abilities, combined with a more expressive pattern language compared to state-of-the-art, capturing mutual exclusive relationship, \ourmethod discovered succinct, informative, and actionable pattern sets that characterize misclassifications of NLP models in two challenging settings, which capture general problems of the model rather than instance specific (local) issues.
It, hence, fills the gap of a robust approach to describe labels in terms of human-interpretable patterns in practice.

While our approach scales already to tens of thousands of features, it makes for engaging future work to scale it even further towards hundreds of thousands of features or to extend the work on characterizing misclassifications incorporating elements of the classifier itself, such as neuron activations~\cite{fischer2021explainn}.

%% file: appendix.tex
\section{Appendix}

\subsection{Pseudocode}
\label{app:pseudocode}

We, here, provide the pseudocode for the function to create candidate patterns (Alg.~\ref{alg:createCandidates}).

\begin{algorithm}[h!]
	\caption{createCandidates} \label{alg:createCandidates}
	\begin{algorithmic}[1]
		\STATE \textbf{Input:} \Data, patterns \Patterns in current \Model, max neighbour distance $K$ 
		\STATE \textbf{Output:} \Candidates, a set of candidate patterns \Patterns
		\STATE  \textit{// Define $\Neighbor(\Item, 0) = \Item$ for simplicity}
		\STATE $\Candidates \gets \{\}$
		\STATE  \textit{// Single item and its neighbours}
		\FOR{$\Item \in \Items$}
		\STATE $A \gets \{\}$
		\FOR{$k \in \{0,\ldots,K\}$}
		\STATE $A \gets A \cup \{\Neighbor(\Item, k)\}$
		\STATE $\Candidates \gets \Candidates \cup \{ \oxor(A) \}$
		\ENDFOR
		\ENDFOR
		\STATE  \textit{// Pairs of items and their neighbours}
		\FOR{$(\Item_1, \Item_2) \in \Items \times \Items$}
		\STATE $A_1 \gets \{\}$
		\FOR{$k_1 \in \{0,\ldots,K\}$}
		\STATE $A_1 \gets A_1 \cup \{\Neighbor(\Item_1, k_1)\}$
		\STATE $A_2 \gets \{\}$
		\FOR{$k_2 \in \{0,\ldots,K\}$}
		\STATE $A_2 \gets A_2 \cup \{\Neighbor(\Item_2, k_2)\}$
		\STATE $\Candidates \gets \Candidates \cup \{ \oland(\oxor(A_1), \oxor(A_2)) \}$
		\ENDFOR
		\ENDFOR
		\ENDFOR
		\STATE  \textit{// Pattern + item and its neighbours}
		\FOR{\Pattern in \Patterns}
		\FOR{$\Item \in \Items$}
		\STATE $A \gets \{\}$
		\FOR{$k \in \{0,\ldots,K\}$}
		\STATE $A \gets A \cup \{\Neighbor(\Item, k)\}$
		\STATE $\Candidates \gets \Candidates \cup \{ \oland(\gamma(\Pattern) \cup \{A\}) \}$
		\ENDFOR
		\ENDFOR
		\ENDFOR
		\STATE \textit{// Pattern + Pattern}
		\FOR{$(\Pattern_1, \Pattern_2) \in \Patterns \times \Patterns$}
		\STATE{$\Candidates \gets \Candidates \cup \{ \oland(\gamma(\Pattern_1) \cup \gamma(\Pattern_2) \}$}
		\ENDFOR
		\STATE  \textit{// see Sections \ref{sec:theory} and \ref{sec:method} for filter criteria}
		\STATE $\Candidates \gets $ \texttt{Filter}$(\Candidates)$
	\end{algorithmic}
\end{algorithm}

\setlength{\tabcolsep}{3.9pt}
\begin{table}[h!]
	\caption{\textit{VQA example patterns.} Our method discovers meaningful and easily interpretable patterns. For the LXMERT dataset, we show a subset of the patterns highlighting different reasons for misclassification along with examples from the corresponding datasets. The full list of retrieved patterns for all methods is given in the Supplementary Material. \label{tab:app_lxmert_examples_vqa}}
	\centering
	\begin{tabular}{p{3.2cm}|l}
		\toprule
		pattern & example \\ \midrule
		\OLAND{How, many} & How many kites are flying? \\
		\OLAND{hanging, from} & What is hanging from a hook? \\
		\OLAND{\OLOR{kind, sort}, of} & What kind of birds are these? \\
		$\oland$($\oxor$(\textit{would, could,} & How would you describe the \\
		\hspace{0.2cm}\textit{might, can})\textit{, you}) & \hspace{0.2cm}decor? \\
		\OLAND{name, of} & What is the name of this \\
		& \hspace{0.2cm}  restaurant? \\
		\textit{number} & What is the pitchers number? \\
		\OLOR{letter, letters} & What letter appears on the box? \\
		$\oland$(\textit{How, much,} & How much does the fruit cost? \\
		\hspace{0.31cm} $\oxor$(\textit{cost,}\textit{costs})) & \\
		\bottomrule
	\end{tabular}
\end{table}

\setlength{\tabcolsep}{3.9pt}
\begin{table*}[h!]
	\caption{\textit{VQA data statistics.} For the two VQA classifiers, we provide general statistics about data dimensions, and for each method the number of discovered patterns ($k=|P|$) or if applicable number of patterns explaining  misclassification ($k^-=|P^-|$), respectively correct classification ($k^+=|P^+|$) and the average pattern length $\overline{|p|}$. }\label{tab:real-data_stats}
	\centering
	\begin{tabular}{l rr rrr rr rr rr rr rr rr rr rr}
		
		\toprule
		&&& \multicolumn{3}{c}{\ourmethod} & \multicolumn{2}{c}{\Tree} & \multicolumn{2}{c}{\textsc{Ripp.}} & \multicolumn{2}{c}{\textsc{Subgr.}} & \multicolumn{2}{c}{\textsc{SpuM.}} & \multicolumn{2}{c}{\textsc{Clas.}}  & \multicolumn{2}{c}{\Grab} & \multicolumn{2}{c}{\textsc{Corta.}} & \multicolumn{2}{c}{\textsc{CSal.}}\\
		\cmidrule(r){4-6}
		\cmidrule(l){7-8}
		\cmidrule(l){9-10}
		\cmidrule(l){11-12}
		\cmidrule(l){13-14}
		\cmidrule(l){15-16}
		\cmidrule(l){17-18}
		\cmidrule(l){19-20}
		\cmidrule(l){21-22}
		Dataset & $|\Items|$& $|\Data|$ & $k^-$ & $k^+$ & $\overline{|p|}$& $k$&$\overline{|p|}$& $k$&$\overline{|p|}$&  $k$ &$\overline{|p|}$& $k$ & $\overline{|p|}$ & $k$ & $\overline{|p|}$ & $k$ & $\overline{|p|}$ & $k$ & $\overline{|p|}$ & $k$ & $\overline{|p|}$\\
		\midrule
		Visual7W & $2429$ & $28032$ & $29$ & $26$ & $3.4$ & $4309$ & $3.6$ & $0$ & $0.0$ & $100$ & $2.3$ & $575$ & $2.9$ & $19$ & $1.3$ & $1$ & $1$ & $15$ & $2.3$ & $10$ & $3.9$ \\
		LXMERT & $5351$ & $25994$ & $41$ & $34$ & $2.7$ & $3371$ & $2.7$ & $3$ & $3.0$ & $100$ & $2.5$ & $951$ & $3.9$ & $36$ & $1.3$ & $1$ & $1$ & $2$ & $3.0$ & $11$ & $4.2$ \\
		\bottomrule
	\end{tabular}
\end{table*}

\subsection{Proof: Order of Items}
\label{app:order_proof}

Here, we provide a proof that the codelength is independent on the order of items in mutual exclusive clauses.
The proof closely follows that of Fischer \& Vreeken~\cite{theory/Fischer20Mexican}.

	Given a clause $\Clause=\oxor(i,j,k)$ with corresponding margins $n_i, n_j, n_k$, it does not matter in which order we transmit
	where the items hold.
	We show that we can flip the item order without changing the cost.
	Assume a new order $P=\oxor(k,i,j)$, then we show
	\begin{align*}
	& \log{n \choose n_i} + \log{n - n_i \choose n_j} + \log{n - n_i - n_j \choose n_k} \\
	\overset{!}{=} & \log{n \choose n_k} + \log{n - n_k \choose n_i} + \log{n - n_i - n_k \choose n_j} \; .
	\end{align*}
	With the definition of the binomial using factorials and standard math,
	adding new terms that add up to 0, we show that the above equation hold.
	\begin{align*}
	&\log\frac{n!}{(n-n_i)!n_i!} + \log\frac{(n-n_i)!}{(n-n_i-n_j)!n_j!} \\ &+ \log\frac{(n-n_i-n_j)!}{(n-n_i-n_j-n_k)!n_k!} \\
	= &\log(n!) - \cancel{\log((n-n_i)!)} - \log(n_i!) + \cancel{\log((n-n_i)!)} \\
	&- \cancel{\log((n-n_i-n_j)!)} - \log(n_j!) + \cancel{\log((n-n_i-n_j)!)}\\
	&-  \log((n-n_i-n_j-n_k)!) - \log(n_k!)\\
	&\underbrace{+ \log((n-n_k)!) - \log((n-n_k)!)}_{=0} \\
	&\underbrace{+ \log((n-n_i-n_k)!) - \log((n-n_i-n_k)!)}_{=0} \\
	=& \log\frac{n!}{(n-n_k)!n_k!} + \log\frac{(n-n_k)!}{(n-n_i-n_k)!n_i!} \\
	&+ \log\frac{(n-n_i-n_k)!}{(n-n_i-n_j-n_k)!n_j!}.
	\end{align*}
	Other permutations and larger clauses follow the same reasoning.

\subsection{Size of the Model Space}
\label{sec:app_modelspace}

We here briefly describe the derivation for the size of the model space.
In particular, the size of the model space is 
\[
	|\Models| = 2^{\sum_{i=1}^{|\Items|} \binom{|\Items|}{i} \times \sum_{j=1}^i \stirling{i}{j}} \;,
\]
where the first term in the summation specifies the number of possible item combinations in a pattern of length $i$, the second term counts the number of possible ways to separate them into $j$ different clauses via the Stirling number of the second kind and the exponent is introduced as a model \Model consists of arbitrary combinations of patterns. The MDL score for such complex model classes does not lend itself for easy-to-exploit structure such as monotonicity. Hence, we resort to an efficient bottom-up search heuristic for discovering good models which we introduce in the next section.

\subsection{Mutual Exclusivity and Word Neighbors}
\label{sec:app_wordneighbors}

For the clauses of mutually exclusive items, we are interested in finding words that are synonyms or that reflect similar concepts, such as \OLOR{color, colour} or \OLOR{could, can}. Research in NLP has proposed various techniques for identifying such pairs including manually created ontologies such as WordNet \cite{approach/Miller95wordnet} or word embeddings that are learned through co-occurrences in text and map words to vector representations. This information about related words can be used to guide the search for mutually exclusive patterns. Using such pretrained embeddings rather than deriving them from the given input data has the advantage that we are independent of the size of the input data set, and receive reliable embeddings, which were trained on very large, domain independent text corpora.

While our approach is independent of the specific method, we have chosen FastText word embeddings trained on CommonCrawl and Wikipedia \cite{embeddings/Grave2018FastText}. In contrast to word ontologies, word embeddings have a broader vocabulary coverage. They also do not impose strict restrictions such as a particular definition of synonyms and instead reflect relatedness concepts learned from the text. FastText embeddings have the additional benefit that they use subword information, removing the issue of out-of-vocabulary words. The word embeddings are independent of the machine learning classifier we study. As measure of relatedness $m$ between two items $\Item_1$, $\Item_2$, we use cosine similarity, i.e. $m = cos(\Embedding(\Item_1), \Embedding(\Item_2))$ where \Embedding\ is the mapping between an item/word and its vector representation. We define $\Neighbor(\Item, k)$ as the $\Item' \in \Items$ for which $m(\Item, \Item')$ is the $k$-highest. Examples for words and their neighbours in FastText embeddings are given in Table \ref{tab:examples_neighbors}.

Based on the information of the embedding, we derive $\oxor$-clauses. For each item \Item, we explore mutual exclusivity in its $1\ldots K$ closest neighbors, i.e. from $\oxor(\Item, \Neighbor(\Item, 1))$ until $\oxor(\Item, \Neighbor(\Item, 1), \ldots, \Neighbor(\Item, K))$ where $K$ is the maximum neighborhood size. For that, we adapt the \texttt{createCandidates} algorithm from Section \ref{sec:method-merge} so that 
whenever we consider merging with an item \Item, we also consider merging with the $\oxor$-clauses containing additionally the  $1,2,\ldots K$ closest neighbours (see Alg. \ref{alg:createCandidates}). 

Since not all words have $K$ neighbors that represent similar words, we additionally filter neighbourhoods such that $\frac{ \bigcap_{\Item} \Selection{\Item}{\Data} }{ \bigcup_{\Item} \Selection{\Item}{\Data} } < a$ and $m(\Item, \Neighbor(\Item, k)) > b_k$  for  all items \Item\ in the clause, i.e. we require that their transactions barely overlap (mutual exclusivity), and that their embeddings are reasonably close.
In all experiments we set $K = 5$, $a = 0.05$ and $b_k$ to the $3$rd quartile of $\{ m(\Item,\Neighbor(\Item, k)) \mid \Item \in \Items \}$.

In the general case for arbitrary labeled data, we could follow the proposal of \citet{theory/Fischer20Mexican} to search for potential XOR structure, which however would lead to a much increased search space and hence computational costs, without any benefits for the specific applications.

\begin{table}
	\begin{tabular}{l l}
		\toprule
		\textbf{Word} & \textbf{5-nearest neighborhood}\\
		\midrule
		\textit{photo} & photograph, photos, picture, pic, pictures \\
		\textit{color} & colour, colors, purple, colored, gray \\
		\textit{can} & could, will, may, might, able \\
		\textit{say} & know, think, tell, mean, want \\
		\bottomrule
	\end{tabular}
	\caption{Words and their nearest neighbors on \emph{Visual7W}.
	}
	\label{tab:examples_neighbors}
\end{table}

\subsection{Complexity}
\label{sec:app_complexity}
While it is common to consider the complexity in terms of the size of the input, the bound it would give -- which is exponential in the number of items as discussed in the theory section -- is neither helpful nor tight considering the discovery of small models. we thus analyze the complexity of \ourmethod in terms of the size of the model. 

Consider \ourmethod finds $k$ conjunctive patterns of maximum length $l$ for a dataset with $m$ items.
Since in every round either a new singleton or pair is generated that belongs to one of the $k$ final patterns, or two existing patterns are merged, the algorithm runs $O(k\,l)$ rounds. In each round, the dominating factor is the candidate generation, out of which there are $O(m)$ potential singletons, $O(m^2)$ pairs, and at maximum $O(k\,l)$ pattern merges, corresponding to the case that all parts of the final patterns exist as singleton patterns in the current round.
Hence, we get a worst case time complexity of $O(k\,l\,(k\,l + m^2))$.

For clauses containing mutual exclusivity, for all practical applications we consider XOR statements of the $c$ closest words in a given embedding, where $c$ is a small constant.
We hence consider $O(m\,c)$ single XOR clauses, $O((m\,c)^2)$ pairs, and at maximum $O(k\,l)$ pattern merges, where again this corresponds to the case that all parts of the final patterns exist as singleton patterns in the current round. Hence we get a worst case time complexity of $O(k\,l\,(k\,l + (m\,c)^2))$.
For the general case, when searching for arbitrary AND and XOR combinations, we refer to Fischer \& Vreeken \cite{theory/Fischer20Mexican}.

\begin{figure*}[h]
	\centering
	\begin{subfigure}{0.24\textwidth}
		\centering
		\ifpdf
		\begin{tikzpicture}
		\begin{axis}[
		eda ybar,
		xtick=data,
		xticklabels={100,200,500,1k,5k,10k},		
		width = 5cm,
		height = 3.5cm,
		xlabel = {\# of items}, 
		ylabel= {F1},
		extra x ticks={0.5,1.5,...,4.5},
		extra x tick labels={},
		every extra x tick/.style={thin, black, major tick length=2pt},
		bar width = 1.1pt,
		xmin=-0.7, xmax=5.5,
		ymin=0, ymax=1,
		legend columns=2,
		x label style 		= {at={(axis description cs:0.5,-0.05)}, anchor=north, font=\scriptsize},
		y label style 		= {at={(axis description cs:-0.1,0.6)}, rotate=90,  anchor=south, font=\scriptsize},
		legend style={nodes={scale=0.9, transform shape}, at={(0.55,0.92)}, anchor=north, row sep=-1.4pt}
		]
		
		\addplot 
		plot [eda errorbar]
		table[x=num_items, y=mijo, y error=mijo_error] {expres/pure-syn-data_evaluation_gen2-4_num-items_hard-f1_22-05-31_modified-xlabel.txt};
		\addplot 
		plot [eda errorbar]
		table[x=num_items, y=tree,y error=tree_error ]{expres/pure-syn-data_evaluation_gen2-4_num-items_hard-f1_22-05-31_modified-xlabel.txt};
		
		\addplot 
		plot [eda errorbar]
		table[x=num_items, y=ripper,y error=ripper_error ]{expres/pure-syn-data_evaluation_gen2-4_num-items_hard-f1_22-05-31_modified-xlabel.txt};

		\addplot
		plot [eda errorbar]
		table[x=num_items,y=subgroup, y error=subgroup_error] {expres/pure-syn-data_evaluation_gen2-4_num-items_hard-f1_22-05-31_modified-xlabel.txt};
		
		\addplot
		plot [eda errorbar]
		table[x=num_items,y=realgrab, y error=realgrab_error] {expres/pure-syn-data_evaluation_gen2-4_num-items_hard-f1_22-05-31_modified-xlabel.txt};
		
		\addplot
		plot [eda errorbar]
		table[x=num_items,y=rulelist, y error=rulelist_error] {expres/pure-syn-data_evaluation_gen2-4_num-items_hard-f1_22-05-31_modified-xlabel.txt};
		
		\addplot
		plot [eda errorbar]
		table[x=num_items,y=spumante, y error=spumante_error] {expres/pure-syn-data_evaluation_gen2-4_num-items_hard-f1_22-05-31_modified-xlabel.txt};
		
		\addplot
		plot [eda errorbar]
		table[x=num_items,y=cortana_negr_02quality, y error=cortana_negr_02quality_error] {expres/pure-syn-data_evaluation_gen2-4_num-items_hard-f1_22-05-31_modified-xlabel.txt};
		
		\addplot
		plot [eda errorbar]
		table[x=num_items,y=csalt, y error=csalt_error] {expres/pure-syn-data_evaluation_gen2-4_num-items_hard-f1_22-05-31_modified-xlabel.txt};
		
		\end{axis}
		\end{tikzpicture}
		\fi
		\caption{Varying number of items\label{fig:pure-syn-data_num-items-f1}}
	\end{subfigure}
	\begin{subfigure}{0.24\textwidth}
		\centering
		\ifpdf
		\begin{tikzpicture}
		\begin{axis}[
		eda ybar,
		xtick=data,
		width = 5cm,
		height = 3.5cm,
		xlabel = {$|\Dlm| / |\Data|$},
		ylabel={},
		extra x ticks={0.25,0.35,...,0.55},
		extra x tick labels={},
		every extra x tick/.style={thin, black, major tick length=2pt},
		enlarge x limits=0.19,
		ymin=0, ymax=1,
		legend columns=2,
		x label style 		= {at={(axis description cs:0.5,-0.05)}, anchor=north, font=\scriptsize},
		y label style 		= {at={(axis description cs:-0.1,0.6)}, rotate=90,  anchor=south, font=\scriptsize},
		legend style={nodes={scale=0.8, transform shape}, at={(0.5,0.9)}, anchor=north, row sep=-1.4pt}
		]
		
		\addplot 
		plot [eda errorbar]
		table[x=l1_percentage, y=mijo, y error=mijo_error] {expres/pure-syn-data_evaluation_gen2-1_l1-percentage_hard-f1_22-05-31.txt};
		
		\addplot 
		plot [eda errorbar]
		table[x=l1_percentage, y=tree,y error=tree_error ]{expres/pure-syn-data_evaluation_gen2-1_l1-percentage_hard-f1_22-05-31.txt};
		
		\addplot 
		plot [eda errorbar]
		table[x=l1_percentage, y=ripper,y error=ripper_error ]{expres/pure-syn-data_evaluation_gen2-1_l1-percentage_hard-f1_22-05-31.txt};
		
		\addplot
		plot [eda errorbar]
		table[x=l1_percentage,y=subgroup, y error=subgroup_error] {expres/pure-syn-data_evaluation_gen2-1_l1-percentage_hard-f1_22-05-31.txt};
		
		\addplot
		plot [eda errorbar]
		table[x=l1_percentage,y=realgrab, y error=realgrab_error] {expres/pure-syn-data_evaluation_gen2-1_l1-percentage_hard-f1_22-05-31.txt};
		
		\addplot
		plot [eda errorbar]
		table[x=l1_percentage,y=rulelist, y error=rulelist_error] {expres/pure-syn-data_evaluation_gen2-1_l1-percentage_hard-f1_22-05-31.txt};
		
		\addplot
		plot [eda errorbar]
		table[x=l1_percentage,y=spumante, y error=spumante_error] {expres/pure-syn-data_evaluation_gen2-1_l1-percentage_hard-f1_22-05-31.txt};
		
		\addplot
		plot [eda errorbar]
		table[x=l1_percentage,y=cortana_negr_02quality, y error=cortana_negr_02quality_error] {expres/pure-syn-data_evaluation_gen2-1_l1-percentage_hard-f1_22-05-31.txt};
		
		\addplot
		plot [eda errorbar]
		table[x=l1_percentage,y=csalt, y error=csalt_error] {expres/pure-syn-data_evaluation_gen2-1_l1-percentage_hard-f1_22-05-31.txt};
		
		\legend{\ourmethod, Tree, \ripper, \textsc{Subgroup}, \Grab, \Classy, \Spumante, \cortana, \CSalt}
		
		\end{axis}
		\end{tikzpicture}
		\fi
		\caption{Varying label ratio \label{fig:pure-syn-data_l1-percentage-f1}}
	\end{subfigure}
	\begin{subfigure}{0.24\textwidth}
		\centering
		\ifpdf
		\begin{tikzpicture}
		\begin{axis}[
		eda ybar,
		xtick=data,
		x dir=reverse,
		width = 5cm,
		height = 3.5cm,
		xlabel = {pattern occurences in $\Dlp$}, 
		ylabel= {},
		extra x ticks={1.05,0.95,...,0.55},
		extra x tick labels={},
		every extra x tick/.style={thin, black, major tick length=2pt},
		enlarge x limits=0.13,
		bar width = 1.4pt,
		ymin=0, ymax=1,
		legend columns=2,
		x label style 		= {at={(axis description cs:0.5,-0.05)}, anchor=north, font=\scriptsize},
		y label style 		= {at={(axis description cs:-0.1,0.5)}, anchor=south, font=\scriptsize},
		legend style={nodes={scale=0.9, transform shape}, at={(0.35,0.9)}, anchor=north, row sep=-1.4pt}
		]
		
		\addplot 
		plot [eda errorbar]
		table[x=shift_value, y=mijo, y error=mijo_error] {expres/pure-syn-data_evaluation_gen2-2_label-shift_hard-f1_22-05-31.txt};
		
		\addplot 
		plot [eda errorbar]
		table[x=shift_value, y=tree,y error=tree_error ]{expres/pure-syn-data_evaluation_gen2-2_label-shift_hard-f1_22-05-31.txt};
		
		\addplot 
		plot [eda errorbar]
		table[x=shift_value, y=ripper,y error=ripper_error ]{expres/pure-syn-data_evaluation_gen2-2_label-shift_hard-f1_22-05-31.txt};
		
		\addplot
		plot [eda errorbar]
		table[x=shift_value,y=subgroup, y error=subgroup_error] {expres/pure-syn-data_evaluation_gen2-2_label-shift_hard-f1_22-05-31.txt};
		
		\addplot
		plot [eda errorbar]
		table[x=shift_value,y=realgrab, y error=realgrab_error] {expres/pure-syn-data_evaluation_gen2-2_label-shift_hard-f1_22-05-31.txt};
		
		\addplot
		plot [eda errorbar]
		table[x=shift_value,y=rulelist, y error=rulelist_error] {expres/pure-syn-data_evaluation_gen2-2_label-shift_hard-f1_22-05-31.txt};
		
		\addplot
		plot [eda errorbar]
		table[x=shift_value,y=spumante, y error=spumante_error] {expres/pure-syn-data_evaluation_gen2-2_label-shift_hard-f1_22-05-31.txt};
		
		\addplot
		plot [eda errorbar]
		table[x=shift_value,y=cortana_negr_02quality, y error=cortana_negr_02quality_error] {expres/pure-syn-data_evaluation_gen2-2_label-shift_hard-f1_22-05-31.txt};
		
		\addplot
		plot [eda errorbar]
		table[x=shift_value,y=csalt, y error=csalt_error] {expres/pure-syn-data_evaluation_gen2-2_label-shift_hard-f1_22-05-31.txt};
		
		\end{axis}
		\end{tikzpicture}
		\fi
		\caption{Varying label shift\label{fig:pure-syn-data_shift-f1}}
	\end{subfigure}
	\begin{subfigure}{0.24\textwidth}
		\centering
		\ifpdf
		\begin{tikzpicture}
		\begin{axis}[
		eda ybar,
		xtick=data,
		xticklabels={0,0.1,0.2,0.5,1},		
		width = 5cm,
		height = 3.5cm,
		xlabel = {\% of flips}, 
		ylabel= {},
		extra x ticks={0.5,1.5,...,3.5},
		extra x tick labels={},
		every extra x tick/.style={thin, black, major tick length=2pt},
		enlarge x limits=0.13,
		bar width = 1.4pt,
		ymin=0, ymax=1,
		legend columns=2,
		x label style 		= {at={(axis description cs:0.5,-0.05)}, anchor=north, font=\scriptsize},
		y label style 		= {at={(axis description cs:-0.1,0.5)}, anchor=south, font=\scriptsize},
		legend style={nodes={scale=0.9, transform shape}, at={(0.5,0.9)}, anchor=north, row sep=-1.4pt}
		]
		
		\addplot 
		plot [eda errorbar]
		table[x=noise_rate, y=mijo, y error=mijo_error] {expres/pure-syn-data_evaluation_gen2-3_background-noise-rate_hard-f1_22-05-31_modified-xlabel.txt};
		
		\addplot 
		plot [eda errorbar]
		table[x=noise_rate, y=tree,y error=tree_error ]{expres/pure-syn-data_evaluation_gen2-3_background-noise-rate_hard-f1_22-05-31_modified-xlabel.txt};
		
		\addplot 
		plot [eda errorbar]
		table[x=noise_rate, y=ripper,y error=ripper_error ]{expres/pure-syn-data_evaluation_gen2-3_background-noise-rate_hard-f1_22-05-31_modified-xlabel.txt};
		
		\addplot
		plot [eda errorbar]
		table[x=noise_rate,y=subgroup, y error=subgroup_error] {expres/pure-syn-data_evaluation_gen2-3_background-noise-rate_hard-f1_22-05-31_modified-xlabel.txt};
		
		\addplot
		plot [eda errorbar]
		table[x=noise_rate,y=realgrab, y error=realgrab_error] {expres/pure-syn-data_evaluation_gen2-3_background-noise-rate_hard-f1_22-05-31_modified-xlabel.txt};
		
		\addplot
		plot [eda errorbar]
		table[x=noise_rate,y=rulelist, y error=rulelist_error] {expres/pure-syn-data_evaluation_gen2-3_background-noise-rate_hard-f1_22-05-31_modified-xlabel.txt};
		
		\addplot
		plot [eda errorbar]
		table[x=noise_rate,y=spumante, y error=spumante_error] {expres/pure-syn-data_evaluation_gen2-3_background-noise-rate_hard-f1_22-05-31_modified-xlabel.txt};
		
		\addplot
		plot [eda errorbar]
		table[x=noise_rate,y=cortana_negr_02quality, y error=cortana_negr_02quality_error] {expres/pure-syn-data_evaluation_gen2-3_background-noise-rate_hard-f1_22-05-31_modified-xlabel.txt};
		
		\addplot
		plot [eda errorbar]
		table[x=noise_rate,y=csalt, y error=csalt] {expres/pure-syn-data_evaluation_gen2-3_background-noise-rate_hard-f1_22-05-31_modified-xlabel.txt};
		
		\end{axis}
		\end{tikzpicture}
		\fi
		\caption{Varying background noise\label{fig:pure-syn-data_background-noise-f1}}
	\end{subfigure}
	\caption{\textit{Synthetic data results (F1 score).} We visualize results on synthetic data with varying number of items (a), label ratio (b), label shift (c) and amount of background noise (d). The results are in terms of F1 score with respect to the ground truth. \label{fig:pure-syn-data_f1}}
\end{figure*}
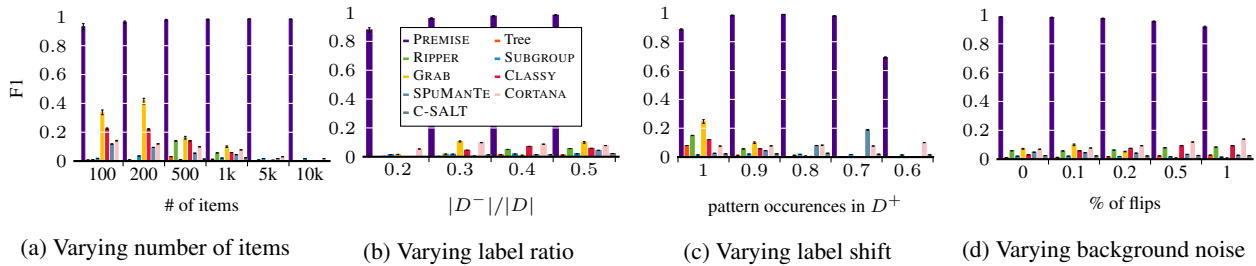

\subsection{Experimental Details}
\label{sec:app_exp_details}
Experiments were performed on an Intel i7-7700 machine with 31GB RAM running Linux.
For the single-threaded C++ implementation of \ourmethod, all synthetic data experiments finished within minutes for the moderately sized data sets, and within hours for the larger datasets with $5$k and $10$k items.
On the VQA datasets \ourmethod finished within 20 minutes and on the NER data within 4 hours. 

For \Tree, patterns are exracted from a decision tree trained on the misclassification data. Each of the tree's inner nodes is a binary decision regarding the presence of an item and a pattern is the conjunctive path from the tree's root to one of its leafs. The model is trained with Gini impurity as decision criterion in the implementation from scikit-learn \cite{decisiontrees/Pedregosa2011scikit}. For \Subgroup, the PySubgroup library is used \cite{subgroup/lemmerich2018pysubgroup} with depth-first search and StandardQF as quality function. The size of the result set and the maximum depth are set to the ground truth for the synthetic data. On the synthetic data, it hence has an advantage over all other approaches which would not hold in a real-world scenario. For the VQA datasets, the values are set to $100$ and $5$ respectively. \Spumante is used with the authors' default parameters, setting its sample size to the dataset size. For \Grab we use the publicly available implementation by the authors, which we tailored for the task at hand by restricting the possible rule-heads to the labels only, but allowing tails over all other items. \cortana \cite{Meeng2011Cortana} is a subgroup discovery tool. It is used by \citet{wouter2014scape} where two numeric labels are expected, one for ground truth and one for prediction probability. We, therefore, split the misclassification label into two labels that disagree if an instance is misclassified. As quality measure, we use negative r and we follow the authors approach of only considering subgroups that cover $1\%$ of the data to prevent overfitting. The maximum depth is set to ground truth for the synthetic data and to $5$ for VQA. The beam width is kept to $100$ and the quality threshold to $0.2$ following the default settings. For \Classy we used the publicly available implementation by the authors as used in the original publication. Minimum support is set to $1$ and maximum rule length to the ground truth for the synthetic data and $5$ for the VQA datasets. For \CSalt~\cite{hess17CSALT}, we use the authors' implementation and the parameters of their test script, i.e. rank increment $10$, $10$k iterations and threshold $0.05$. We set the maximum rank to ground truth for the synthetic data and to $100$ on the VQA data.

For Visual7W and LXMERT, we use the published, pretrained models by the corresponding authors. For LXMERT, the minival version of the development set is used. For the LSTM+CNN+CRF classifier \cite{ner/ma2016lstmcnncrf} for NER, we follow the specific set-up from \citet{ner/hedderich2020african} with English FastText embeddings. %
OntoNotes was split and preprocessed using the script from \url{https://github.com/yuchenlin/OntoNotes-5.0-NER-BIO}.
The fine-tuning data consists of 240 instances/sentences as two patterns did not match any training data. Fine-tuning on the additional data is performed for 30 epochs. As labels, the intersection between CoNLL03 and OntoNotes is used (PER, LOC, ORG) in the BIO2 format.

\subsection{F1 Metric}
\label{sec:app_f1}
A standard metric to evaluate success of a model is the F1 score -- the harmonic mean between precision and recall -- which for discovered pattern set $P_d$ and ground truth pattern set $P_g$ is defined as $\text{F1}(P_d, P_g) = |P_d \cap P_g|/\left(|P_d \cap P_g| + \frac{1}{2}|P_d \symdiff P_g|\right)$, where $\symdiff$ is the symmetric difference between two sets.
As competitors only recover fragments of patterns and hence they obtain very low F1 scores, we instead report a soft F1 score that rewards also fragments. We define it as harmonic mean between a soft precision and a soft recall:
\begin{align}
 \text{SoftPrec}(P_d, P_g)&=\sum_{p_d \in P_d} \argmax_{p_g \in P_g} \frac{|p_d \cap p_g|}{|p_g|}\;,\\
 \text{SoftRec}(P_d, P_g)&=\sum_{p_g \in P_g} \argmax_{p_d \in P_d} \frac{|p_d \cap p_g|}{|p_g|}\;,\\
  \text{F1}(P_d, P_g) &= \frac{2*\text{SoftPrec}*\text{SoftRec}}{\text{SoftPrec} + \text{SoftRec}}\;.
\end{align}
Results with the original F1 score are given in Figure \ref{fig:pure-syn-data_f1}.

\subsection{Synthetic Text Data Experiments}
\label{sec:app_syn_text}

To obtain a synthetic data set with similar item/token distributions as natural language text, we derive transactions/instances from the around 3.4k sentences in the development set of the PennTreebank Corpus. %
In particular, we draw $12$ distinct patterns, for each pattern choosing items from the vocabulary tokens at random.
To ensure that we introduce only new patterns into the data, we verify that none of the items in the patterns co-occur in the original data.
We then insert each pattern into a random subset of the PennTreebank instances, where the number of instances to be covered is drawn from a normal $\Normal(150,20)$. The data contains $6$k unique items.
To evaluate settings typical for classification, we then vary two types of noise. \textit{Shift noise} indicates the percentage of instances with a pattern that are actually labeled as misclassifications, lower values mean that the model is still able to predict correctly in some of the instances -- e.g. because a network leverages additional information in the data.
The second type of noise is labeling instances as misclassification although there is no pattern occurrence -- i.e. non-systematic errors -- which we refer to as \textit{label noise}.
For all samples with pattern occurences, we label a fraction of those as misclassification according to the \textit{shift noise}, and then introduce \textit{label noise}.

\paragraph{Experimental setups} We generate four different sets of experiments. In the first set, we introduce conjunctive patterns varying pattern length of the introduced patterns between 1 and 8 without noise.
In the second set of experiments we vary the amount of \textit{shift noise}, introducing shifts of $\{0.6,0.7,0.8,0.9,1\}$, and choosing pattern length uniformly in 1 to 5.
In the third set we instead change the amount \textit{label noise}, varying in $\{0, 0.05, 0.1, 0.15, 0.2\}$.
In the fourth set of experiments, we introduce patterns consisting of conjunctions of mutual exclusive itemsets. The number of clauses per pattern and the number of items for each clause is chosen uniformly at random between $1$ and $5$. A pattern is only added to an instance if this would not break the mutual exclusivity assumptions of all patterns. For the word neighborhoods, items in the same clause obtain embeddings located around a randomly chosen centroid. All other items obtain random embeddings.
We repeat all experiments 10 times and report the F1 score -- the harmonic mean between precision and recall -- as average across repetitions.

\paragraph{Results}
For the first experiment set (Fig.~\ref{fig:syn-data_onlyand_pattern-length}) of varying pattern length, we observe that \Subgroup and \cortana are able to retrieve short patterns well, failing however to discover any larger patterns, instead retrieving large sets of redundant patterns. Decision trees perform similarly due to overfitting, finding a plethora of highly redundant patterns.
\Spumante, which although based on statistical testing, consistently finds thousands of redundant patterns, performing worst of all in this regard. 
The rule set miner \Grab recovers small patterns well, it performs however much poorer in retrieving patterns of larger size.
\ourmethod is the only approach to consistently recover the ground truth in all data sets.

For both noise experiments, visualized in Fig.~\ref{fig:syn-data_onlyand_shift} and \ref{fig:syn-data_onlyand_misclass-without-pattern}, 
the tree based method completely breaks down already for moderate amounts of noise.
\Subgroup and \Spumante both perform consistently bad with F1 scores below $.2$.
Out of the existing approaches, only \Grab is able to recover the ground truth well. \ourmethod outperforms all existing methods in each of our noise experiments, achieving consistently high F1 scores beyond $.92$.

As ablation, we ran \ourmethod without the noise filter introduced in Section \ref{sec:fisher}. A score around $0.3-0.4$ is achieved, which is better than most baselines, but much lower than the score achieved with the complete \ourmethod method.

Since most baselines do not support discovering mutual exclusivity or proved to fail in the more simple setup of conjunctions, we only evaluate our proposed method on the fourth set of experiments. 
We observe that \ourmethod is still able to retrieve patterns even in this challenging setup of complex clauses, with F1 scores close to $.9$, and is able to discover clauses in the presence of noise (Fig.~\ref{fig:syn-data_and+or}).